%% file: main.tex
\documentclass[nohyperref]{article}

\usepackage{microtype}
\usepackage{graphicx}
\usepackage{booktabs} 

\usepackage[style=base]{caption}
\usepackage{hyperref}

\usepackage[accepted]{icml2022}

\input{math_commands.tex}

\usepackage{tikz}
\usepackage{algorithm}
\usepackage{algpseudocode}
\usepackage[normalem]{ulem}
\usepackage{bbm}
\usepackage{upgreek}
\usepackage{tikz}
\usepackage{hyperref}
\usepackage{url}
\usepackage{microtype}
\usepackage{enumitem}
\usepackage{amsmath}
\usepackage{amssymb}
\usepackage{xcolor}
\usepackage{subcaption}
\usepackage{mathtools}
\usepackage{graphicx}

\input{leons_defs}

\setlist[itemize]{itemsep=0.0em, parsep=0.2em, topsep=0pt}

\graphicspath{ {./images/} }

\newtheorem{theorem}{Theorem}
\newtheorem{lemma}{Lemma}
\def\Atechnical{%
    \bm{A}_{\mathcal{X},\rho,\mathbb{P}_X}^{\text{technical}}
}
\def\Abesicovitch{%
    \bm{A}_{\eta}^{\text{Besicovitch}}
}
\def\Alipschitz{%
    \bm{A}_{\eta}^{\text{Lipschitz}(\vartheta_{\max})}
}
\def\dnn{\hat \eta}
\newcommand{\dnnlong}{\hat \eta_{\text{DNNR}}}
\def\P{\mathbb{P}}
\def\B{B}
\def\Bxh{\B_{x,h}}

\begin{document}

\twocolumn[

\icmltitle{DNNR: Differential Nearest Neighbors Regression}



\icmlsetsymbol{equal}{*}

\begin{icmlauthorlist}
\icmlauthor{Youssef Nader}{equal,fu}
\icmlauthor{Leon Sixt}{equal,fu}
\icmlauthor{Tim Landgraf}{fu}
\end{icmlauthorlist}

\icmlaffiliation{fu}{Department of Computer Science, Freie Universität Berlin, Germany}

\icmlcorrespondingauthor{Youssef Nader}{youssef.nader@fu-berlin.de}
\icmlcorrespondingauthor{Leon Sixt}{leon.sixt@fu-berlin.de}
\icmlcorrespondingauthor{Tim Landgraf}{tim.landgraf@fu-berlin.de}

\icmlkeywords{Machine Learning, KNN, ICML}

\vskip 0.3in
]

\printAffiliationsAndNotice{\icmlEqualContribution} 

\begin{abstract}
K-nearest neighbors (KNN) is one of the earliest and most established algorithms in machine learning. For regression tasks, KNN averages the targets within a neighborhood which poses a number of challenges: the neighborhood definition is crucial for the predictive performance as neighbors might be selected based on uninformative features, and averaging does not account for how the function changes locally. We propose a novel method called Differential Nearest Neighbors Regression (DNNR) that addresses both issues simultaneously: during training, DNNR estimates local gradients to scale the features; during inference, it performs an n-th order Taylor approximation using estimated gradients. In a large-scale evaluation on over 250 datasets, we find that DNNR performs comparably to state-of-the-art gradient boosting methods and MLPs while maintaining the simplicity and transparency of KNN. This allows us to derive theoretical error bounds and inspect failures. In times that call for transparency of ML models, DNNR provides a good balance between performance and interpretability.\setcounter{footnote}{1}\footnote{For code, see supplementary material.}
\end{abstract}

\section{Introduction}

K-nearest neighbors (KNN) is an early machine learning algorithm  \cite{cover1967nearest}
and a prototypical example for a transparent algorithm.
Transparency means that a model's decision can be explained by inspection of its parts. 
KNN's transparency follows from its simplicity: it can be expressed in simple terms as "the system averages the targets of the most similar points to the query". 
At the same time, the algorithm's simplicity makes it amenable for theoretical analysis, such as obtaining bounds on KNN's prediction error \cite{chaudhuri2014}.

\begin{figure*}[t]
    \centering
    \begin{subfigure}[t]{0.45\linewidth}
    \includegraphics{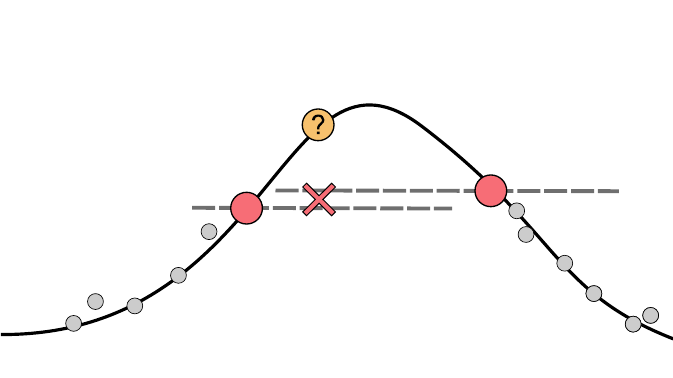}
    \caption{KNN}
    \end{subfigure}
    \begin{subfigure}[t]{0.45\linewidth}
    \includegraphics{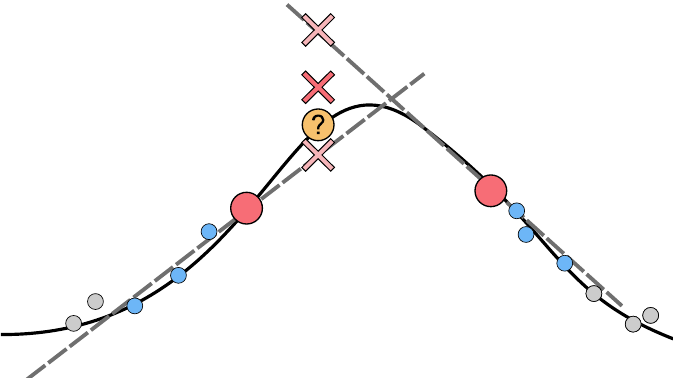}
    \caption{DNNR}
    \end{subfigure}
    \caption{
        \textbf{(a)}
        An illustration of KNN regression. 
        To predict a value for a query (circle with 
        question mark),
        the target values of the nearest points (red circles) are averaged. KNN's prediction is marked by the red cross.
        The other data points (gray circles) are not used for prediction.
        \textbf{(b)} Similar illustration of DNNR.
        The local gradient (gray dashed line) is estimated for each neighbor and a 
        target value is interpolated linearly (light red crosses). The final prediction (red cross) is the average of these interpolated values.
    }
    \label{fig:figure1}
    \vspace{-0.5cm}
\end{figure*}

However, KNN's predictive performance is limited. Most works aiming to improve KNN primarily focused on the selection of the neighbors, the distance metric, and the number of nearest neighbors $k$
\cite{Wettschereck93,weinberger2007metric,entropy_distance}.
KNN's averaging scheme assumes that the target variable's changes are independent of those in the input features. Here, we introduce \emph{Differential Nearest Neighbor Regression} (DNNR) to make use of that very gradient information.
For each neighbor of a query point, we estimate the gradient of the function, and then -- instead of averaging the targets -- we average a Taylor approximation. KNN can then be seen as a zero-order Taylor approximation, while DNNR uses higher orders of the Taylor's theorem. 
A visual summary of the differences between KNN regression and DNNR can be found in Figure \ref{fig:figure1}. 

In a theoretical analysis,
we derived a bound on the point-wise error of DNNR in relation to parameters such as the number of training points and the neighborhood size and found that the error bound favors DNNR over KNN.
In an empirical evaluation on over 250 different datasets, we confirmed that DNNR outperforms KNN regression and performs on par with gradient boosting methods. 
An ablation study then confirmed that both the gradient-based prediction and the feature scaling contribute to the performance gains.
Using a synthetic dataset generated with known underlying ground truth, we simulated the error bound and found that DNNR requires fewer training points than KNN.
Furthermore, we present an investigation on a DNNR failure and showcase how the model's transparency can be used in such an analysis.

The regulation of Machine Learning algorithms in high-risk applications is being discussed globally or already under preparation \cite{eu2021proposal}. 
We contribute to the overall goal of transparent and safer ML models in the following ways:
\begin{itemize}
    \setlength\itemsep{0.0em}
    \item We propose a new regression method (DNNR) that performs on par with state-of-the-art algorithms;
    \item DNNR is theoretically grounded: we provide a proof to bound DNNR's point-wise error 
        (Theorem \ref{theorem:DNNR_pointwise_error}) and validate its usefulness empirically;
    \item An extensive evaluation against 11 methods on a set of 8 regression datasets, the PMLB benchmark (133 datasets), and Feynman symbolic regression (119 datasets); 
    \item 
    We provide detailed analyses to understand DNNR's performance (ablation study; impact of data properties) and transparency (inspection of failure cases).  
\end{itemize}

\section{Related Work}

\fakeparagraph{KNN} is a non-parametric model based on a simple voting decision rule where the target of a given point is predicted by averaging the targets of neighboring samples \cite{cover1967nearest}. For an introduction to KNN, we refer the reader to \cite{Chen2018ExplainingTS}.

Numerous methods have been proposed to improve this simple decision rule. \cite{Kulkarni1995,chaudhuri2014} investigated the KNN convergence rate under different sampling conditions while \cite{Balsubramani2019AnAN} and \cite{Wettschereck93} proposed different methods for an adaptive choice of $k$.   

Although the choice of the number of neighbors is critical, it is not the only factor governing KNN performance. A large subset of the KNN literature proposed techniques for the choice of the distance metric that defines the neighborhood. \cite{entropy_distance} introduced an entropy-based distance metric, and \cite{WANG2007207} proposed an adaptive distance. Metric learning methods propose data-driven learning mechanisms for the distance function \cite{weinberger2007metric,large_margin_knn,wang2018}. A similar approach changes the data representation upon which the distance function operates via feature weighting or feature selection \cite{Aha1998,Vivencio2007FeatureweightedKN}. 
\cite{knn_survey} provides a more comprehensive overview of the different KNN techniques.
However, all these methods do not change how the prediction is being performed -- all use an averaging scheme of the targets that effectively does not account for how the function changes within the local neighborhood. 

A method that uses local changes is 
local linear regression (LL) \cite{fan1992design}. 
Similarly to KNN, local linear regression selects $k$-nearest neighbors and then fits a hyperplane locally. This differs from our proposed method as we fit the gradient for all nearest neighbors separately. The single hyperplane of LL regression assumes an identical gradient for each neighbor. We show results for LL regression in the quantitative evaluation.

\fakeparagraph{Gradient Approximation}
Estimating the gradient from data has been studied for various reasons, including variable selection and dimensionality reduction \cite{structuremarian2001,mukherjee2006covariance}. 
Several non-parametric methods exist to estimate the gradient from  data \cite{localpolyfan1996,debarbanter2013}, and bounds of convergence already exist for some techniques \cite{gradient_approx_d_free,Turner2010ErrorBF}.  
An error bound on L1-penalized gradient approximation using KNN is derived in 
\cite{benshabat2020deepfit} using local gradient approximation for 3D model reconstruction by fitting truncated jets in the KNN neighborhoods.
\cite{ausset2021nearest}.
The work also applied the estimated gradient to variable selection and gradient optimization problems, but did not utilize it to improve the prediction. 
As we will present in detail, our method combines non-parametric gradient estimation using KNN, Taylor expansion, and feature scaling for regression modeling. To our surprise, 
this combination was neither explored theoretically nor empirically before.

\section{Method}

\begin{algorithm}[t]
    \caption{Pseudocode of DNNR's prediction for a query point $X$. 
    The feature scaling is omitted in this pseudocode. The OLS function solves an ordinary-least-squares problem.
    }
    \label{alg:DNNR_prediction}
    \begin{algorithmic}
        \Require a query point $x$, train data $\{(X_i, Y_i)\}$, nearest neighbors $k$, 
        nearest neighbors for gradient estimation $k'$, range for target value $y_{\min}, y_{\max}$:
        \State $M \gets $ nn(x, k)
        \State \# $M$ contains the indices of the $k$ nearest neighbors
        \State predictions = []
        \For{ each neighbor index  m $\in M$}
            \State A $  \gets $ nn($X_m, k'$)   
            \State \# {
                $A$ contains indices of the $k'$ neighbors for $X_m$ 
            }
            \State $\Delta \bm{X} \gets {X}_{A} - x_m $ 
            \State $\Delta \bm{Y} \gets {Y}_{A} - y_m $
            \State $\hat \gamma_m \gets \text{OLS}(\Delta \bm{X}, \Delta \bm{Y})$
            \State \# { $\hat \gamma_m$ approximates the gradient}
            \State $\hat y_m \gets y_m + \hat \gamma (x_m - x)$
            \State predictions.append($\hat y_m$)
        \EndFor
        \State $\hat y$ = mean(predictions)
        \State \Return clip($\hat y, y_{\min}, y_{\max}$)
    \end{algorithmic}
\end{algorithm}

\fakeparagraph{Notation}
We will consider the typical supervised regression problem. The training data is 
given as a set of $n$ tuples of data points and target values $\{(X_1, Y_1), \ldots, (X_n, Y_n)\}$. 
We will denote the expected target value by  $\eta(x) = \E[Y|X = x]$.
A ball with radius $r$ around $x$ is given by $ B_{x, r}$. 
For a ball around $x$ with exactly $k$ training points, we will
use a $\#$ sign as in  $ B_{\vx, \#k}$. We summarize our notation in the Appendix Table \ref{tab:notation}.

\fakeparagraph{DNNR}
Vanilla KNN predicts the target of a given datapoint $x$ by averaging the targets of nearby training points:
\begin{equation}
    \eta_{\text{KNN}}(x) = \frac{1}{k} \sum_{
    \mathclap{
    X_m \in B_{x, \#k}
    }} Y_m.
\end{equation}
This simple averaging scheme can be seen as  
a zeroth-term Taylor expansion around the inference point.
If we would know the gradient of the function $\eta$, we could
easily extend it to a first degree Taylor expansion: 
\begin{equation}
    \label{eq:eta_true_grad}
    \eta_{\text{known} \nabla}(x) = \frac{1}{k} \!
    \sum_{
        \mathclap{
        X_m \in B_{x, \#k}
    }
    }
        \left(
            Y_m + \nabla \eta(X_m) (x - X_m) 
        \right).
\end{equation}
Of course, we only have access to the training data, not the underlying target function.
Therefore, we approximate the gradient using nearby points.
We can approximate $\nabla \eta(X_m)$ by solving a least-squares problem:
\begin{equation}
    \label{eq:grad_approximation_text}
    \hat \gamma_m = \arg \min_{\gamma_m} || 
        A \gamma_{m} - q
    ||, 
\end{equation}
where $\hat \gamma_{m} \in \R^d$ is the estimated gradient at point $X_m$,
$A \in \R^{k'\times d}$ contains the normalized differences $(X_i-X_m)/h_i$
as row vectors, $h_i = ||X_i-X_m||$,
$k'$ is the number of points used to approximate the gradient,
$i$ indexes these $k'$ nearby points around $X_m$ ,
and $q \in R^{k'}$ denotes the differences in the labels $q = (Y_i - Y_m) / h_i$.
The result $\hat \gamma_m$ can then substitute the real gradient in equation \eqref{eq:eta_true_grad} to yield the DNNR approximation:
\begin{equation}
    \eta_{\text{DNNR}}(x) = \frac{1}{k} 
    \sum_{
        \mathclap{
        X_m \in B_{x, \#k}
    }}
        \left(
            Y_m + \hat \gamma_m (x - X_m) 
        \right).
\end{equation}
Using the approximate Taylor expansion, we aim to improve the prediction performance by also utilizing the local change in the function.
The pseudocode of DNNR is listed in Algorithm \ref{alg:DNNR_prediction}.
The algorithm can also be extended easily to higher-orders of a Taylor expansion. In the evaluation, we will report results using the diagonal of the Hessian, i.e. the elements corresponding to $\frac{
\partial^2 \eta}{\partial^2 x_i}$.

\paragraph{Feature Weighting}
Using an isotropic distance weighs each dimension equally. 
This way, a neighbor may be picked based on irrelevant dimensions.
A simple improvement is to scale the feature dimensions as done in previous work \cite{weinberger2007metric}.
We use a common approach for a distance metric $d$ using a diagonal matrix $W$ to scale each dimension:
\begin{equation}
    d(x_i, x_j) = (x_i-x_j)^T W (x_i,x_j)
\end{equation}
DNNR's predictions are differentiable w.r.t. the input dimensions, so a loss can be backpropagated to the scaling matrix $W$ and optimized using gradient decent.
Inspired by the Taylor theorem, we require that nearby points predict each other well while 
predictions of far points may come with a larger error.
Therefore, the loss enforces a correlation between distance and the prediction error:
\begin{equation}
\label{eq:feature_weighting_optimization}
\begin{split}
W^* = \arg \min_{W}
    \sum_{i, j \in I}
        \text{corr}\big(
            &  d(X_i,X_j), \\[-10pt]
            & \left| 
                    Y_i -  \dnnlong(X_{\text{nn}(i, k')})
            \right| \big),
\end{split}
\end{equation}
where $I$ is an index set of nearby points, and $\text{corr}$ denotes the Pearson correlation coefficient.

At first sight, an alternative might have been minimizing the prediction's mean squared error (MSE). However, 
minimizing the MSE would not alter the scaling, 
as the  prediction is scale-invariant, i.e. downscaling a dimension will increase the gradient, and the prediction will stay the same. 
Therefore, a spatial inductive bias in equation \ref{eq:feature_weighting_optimization} is needed.

\section{Theoretical Analysis}
\label{sec:theory}

We focus on the point-wise error estimate of DNNR vs. KNN regression. 
The proof contains two parts: the approximation error of the gradient and the point-wise prediction error.
In Appendix \ref{appendix:local_grad_approx}, we show that:
\begin{lemma}
\label{lemma:grad_estimation}
Let $f: D \subset \R^d \to \R $ be of class $C^\mu$,  $a \in D$, and $\mathcal{B}(a) \subset D$
be some neighborhood of a.
Suppose that
around point $a$ we have $m$ neighboring points $v_k$, $ k = 1, \ldots, m$ 
with a, $v_1, \ldots, v_m \in \mathcal{B}(a) \subset D$.
Suppose further that all $\mu$-th order derivatives are Lipschitz, $i \in 1 \ldots \mu$:
$ \frac
    {\partial^i f}
    {\partial a_1^{l_1} \ldots \partial a_d^{l_d}}
    \in Lip_{\vartheta_{i}}(\mathcal{B}(a))
$
where $l_1 + \ldots + l_d = i$
and we approximate the gradient locally at $a$
by $\hat \gamma = E_1 \hat \omega$ via the least-squares solution
$\hat \omega = \arg \min_{\omega \in \R^d} || A\omega - q||$,
where
\begin{equation}
    A\!=\!\begin{pmatrix}
        \nu_1^T & \nu_{1*}^T \\
        \nu_2^T & \nu_{2*}^T \\
        \vdots               \\
        \nu_m^T & \nu_{m*}^T \\
    \end{pmatrix},
    \quad 
    q\!=\!\begin{pmatrix}
    \!
        \frac{f(a + h_1 \nu_1) - f(a)}{h_1} \\
        \frac{f(a + h_2 \nu_2) - f(a)}{h_2} \\
        \vdots \\
        \frac{f(a + h_m \nu_m) - f(a)}{h_m} \\
    \end{pmatrix},
\end{equation}
$A\!\in\!\R^{m\times p}$, 
$q\!\in\!\R^{m}$,
$E_1 = \begin{pmatrix}
    I_d & 0 
\end{pmatrix}
\in \R^{d\times p}
$, 
$ h_k = || v_k - a || $
with 
$ h_k \nu_k = v_k - a$;
$ \nu_k^T = (\nu_{1_k}, \ldots, \nu_{d_k}) $,
$ \nu_{k*}^T $ denotes higher-order terms
$ \nu_{k*}^T = 
\begin{pmatrix}
    \frac{h_k^{\mu'-1}}{\mu'!} 
    \binom{\mu'}{l_1, \ldots, l_d} 
    \prod_{i_1}^d \nu_i^{k_i}
\end{pmatrix}_{2 \le \mu' \le \mu, l_1 + \ldots l_d = \mu'}
$
and $
p = \sum_{i=1}^{\mu} \frac{(d+\mu)!}{d!}
$.
Then a bound on the error in the least-squares gradient estimate is given by: \begin{equation}
    || \nabla f(a) - \hat \gamma ||_2 \le \frac{\vartheta_{\max} h_{\max}^\mu}{\sigma_1 (\mu+1)! } \sqrt{\sum_{i=1}^m ||\nu_i||_1^{2\mu}},
\end{equation}
where $\sigma_1$ is the smallest singular value of A, which is assumed to have $\text{rank}(A) = p$,
$\vartheta_{\max} = \max_{i \in 1\ldots k} \vartheta_i$
and $h_{\max} = \max_{1\le k \le m} h_k$.
\end{lemma}

This lemma extends a result from 
the two-dimensional case \cite{Turner2010ErrorBF}.
The gradient approximation depends on the Lipschitz constant $\vartheta_{\max}$, the distance to the neighbors $h_{\max}$, and
the smallest singular value $\sigma_1$ of the normed differences $A$.
The lemma also shows that by accounting for higher-order terms, e.g. picking $\mu > 1$, the gradient approximation can be made more accurate.

The following theorem is based on  
Theorem 3.3.1 in \cite{Chen2018ExplainingTS}.
We built on the same assumptions
except requiring Lipschitz instead of Hölder continuity. 

\fakeparagraph{Technical Assumptions ($\Atechnical$):}
    The feature space $\mathcal{X}$ and distance $\rho$
    form a separable metric space.
    The feature distribution $\mathbb{P}_X$ is a Borel probability measure.

\fakeparagraph{Assumption Besicovitch($\Abesicovitch$):}
    The regression function $\eta$ satisfied the Besicovitch 
    condition if
    $\lim_{r \downarrow 0} \E[Y | X \in \B_{x, r}]\!=\!\eta(x)$
    for $x$ almost everywhere w.r.t. $\mathbb{P}_X$.

\fakeparagraph{Assumption Lipschitz ($\Alipschitz$):}
    The regression function $\eta$ is Lipschitz
    continuous with parameter $\vartheta_{\max}$ if
    $|\eta(x) - \eta(x') | \le \vartheta_{\max} \rho(x, x') $
    for all $x, x' \in \mathcal{X}$.

Using Lemma \ref{lemma:grad_estimation}, we prove the following theorem in Appendix \ref{appendix:local_grad_approx}:
\begin{theorem}(DNNR pointwise error)
\label{theorem:DNNR_pointwise_error}
Under assumptions
$\Atechnical$ and $\Abesicovitch$,  
let 
$x \in \text{supp}(\P_X)$ be a feature vector, 
$\varepsilon > 0$ be an error tolerance 
in estimating the expected label $\eta(x) = \E[Y| X = x]$,
and $\delta \in (0, 1)$ be a probability tolerance.
Suppose that $Y \in [y_{\min}, y_{\max}]$ for some 
constants $y_{\min}$ and $y_{\max}$.
There exists a threshold distance $h_\text{DNNR}^* \in (0, \inf)$
such that for any smaller distance $h \in (0, h^*)$,
if the number of training points $n$ satisfies:
\begin{equation}
    n \ge \frac{8}{\P_X(\Bxh)} \log \frac{2}{\delta},
\end{equation}
and the number of nearest neighbors satisfies
\begin{equation}
        \frac
            {2 (y_{\max} - y_{\min})^2 }
            {\varepsilon^2}
        \log \frac{4}{\delta} 
    \le 
        k
    \le 
        \frac{1}{2}
        n \P_X(\Bxh),
\end{equation}
then with probability at least $ 1 - \delta $ over randomness
in sampling the training data, DNNR regression at point $x$ has error
\begin{equation}
    |\dnnlong(x) - \eta(x)| \le \varepsilon.
\end{equation}
If we know that $\Alipschitz$ also holds, we can pick $h_{\text{DNNR}}^*$ as:
\begin{equation}
\label{eq:h_star_dnn}
     h_{\text{DNNR}}^* = \sqrt{
        \frac
            {\varepsilon }
            {\vartheta_{\max}
            \left(
                1 
                + 
                \tau
            \right)
            }
    },
\end{equation}
where $
    \tau\!=\!\E \left[
            \frac{
                  \sqrt{\sum_{i=1}^m ||\nu_i||_1^{2\mu}}
            }
                  {\sigma_1 } 
            \:\big|\:  X \in  \Bxh 
        \right] 
$.
$\nu$, $\sigma_1$,  and 
$\vartheta_{\max}$ are defined as in Lemma \ref{lemma:grad_estimation}.
\end{theorem}

The theorem states that we can bound the point-wise error locally with a high probability given that the conditions on $n$ and $k$ are fulfilled. 
Theorem 3.3.1 from \cite{Chen2018ExplainingTS} provides a 
KNN regression point-wise error bound (their theorem is in turn based on \cite{chaudhuri2014}).
For KNN, the restriction on the maximum distance  $h_{\text{KNN}}^*=\frac{\varepsilon}{2 \vartheta_{\max}} $, while the other conditions on sample size and probability are identical. 
To compare DNNR and KNN, 
it is beneficial to solve for the error tolerance $\varepsilon$.
\begin{equation}
    \label{eq:dnnr_error_tolerance}
    \varepsilon_{\text{DNNR}} =
    h_{\text{DNNR}}^2 
    {
        \vartheta_{\max}
        \left(1 + \tau \right)
    },
\end{equation}
and for KNN:
\begin{equation}
    \label{eq:knn_error_tolerance}
    \varepsilon_{\text{KNN}} = 2 \vartheta_{\max} h_{\text{KNN}}.
\end{equation}
The influence of the different variables is as follows:
\begin{itemize}
    \item Both depend on the 
        Lipschitz constant  $\vartheta_{\max}$ linearly.
    \item Distance to nearest neighbors: $h_{\max}$ vs $h_{\max}^2$.
        For DNNR, the error decreases quadratically in $h_{\max}$. When $h_{\max}$ becomes small, $h_{\max}^2$ will be even smaller.
    \item For DNNR, $\tau$ represents the error in estimating the gradient.
        As long $\tau < \frac{2}{h_{\max}} - 1$, the error tolerance of DNNR will be lower than for KNN.
        As $\tau \propto \frac{1}{\sigma_1} $, an ill-conditioned matrix $A$ might increase DNNR's error. 
\end{itemize}

\section{Experiments}
\label{sec:experiments}

We compared DNNR against other methods, including state-of-the-art gradient boosting methods. 
First, we discuss which baselines we compared against and the general experimental setup. 
Then, we present large-scale quantitative evaluations followed by an ablation study and further qualitative analyses.

\subsection{Setup}

We compared DNNR against state-of-the-art boosting methods
(CatBoost, 
XGBoost, 
Gradient Boosted Trees \cite{Dorogush2018CatBoostGB,xgboost}) 
and classical methods such as (KNN, multi-layer perceptron (MLP), 
Random Forest). We also included KNN-Scaled, which uses DNNR's feature weighting but KNN's averaging scheme.

We used standard-scaling for all datasets 
(zero mean, unit variance per dimension). Each model was optimized using a grid search over multiple parameters, and the models were refit using their best parameters on the validation data before test inference, 
except for 
TabNet \cite{Arik_Pfister_2021} (for details see Appendix 
\ref{appendix:hyperparams}).
We ensured that each method had a comparable search space.
Approximately 4.1k CPU hours were used to run the experiments. 
For the larger benchmarks (Feynman and PMLB), we followed the setup in 
\cite{cava2021contemporary}.
Our baselines report similar or slightly better performance than on SRBench\footnote{
See results here: https://cavalab.org/srbench/blackbox/
}. 

\subsection{Quantitative Experiments}
\label{subsec:quantitative_experiments}

\begin{table*}[t]
    \vspace{-0.4cm}
    \centering
    \caption{
        The MSE on eight regression datasets averaged over 10-folds.
        The best performing values are marked as bold.
        The standard deviations are given after the ± signs. For the Sarcos dataset, we only evaluate on the given test set.
        }
    \label{tab:results_datasets}
    \small
    \setlength{\tabcolsep}{0.47em} 
\begin{tabular}{lllllllll}
\toprule
{} &       Protein &  CO$^2$Emission &   California &          Yacht &       Airfoil &       Concrete &   NOxEmission & Sarcos \\

\midrule
CatBoost               &  11.82 ± 0.33 &  1.11 ± 0.40 &  \textbf{0.19 ± 0.01} &    \textbf{0.25 ± 0.25} &   \textbf{1.26 ± 0.24} &   \textbf{14.00 ± 5.35} &  14.65 ± 1.00 &  1.313 \\
Grad. B. Trees          &  12.15 ± 0.41 &  1.25 ± 0.44 &  0.20 ± 0.01 &    0.33 ± 0.15 &   1.76 ± 0.51 &   16.03 ± 6.13 &  15.60 ± 1.00 &  1.813 \\
LGBM                   &  12.74 ± 0.35 &  1.17 ± 0.41 &  0.19 ± 0.01 &    6.86 ± 9.10 &   2.03 ± 0.38 &   16.10 ± 5.11 &  15.66 ± 1.00 &  1.613 \\
MLP                    &  14.34 ± 0.44 &  1.23 ± 0.47 &  0.35 ± 0.26 &   4.34 ± 10.21 &  10.17 ± 3.01 &  35.27 ± 12.17 &  19.91 ± 1.82 &  1.277 \\
Rand. Forest          &  11.80 ± 0.25 &  1.12 ± 0.43 &  0.23 ± 0.02 &    1.13 ± 0.73 &   2.81 ± 0.66 &   22.75 ± 5.64 &  16.35 ± 1.03 &  2.264 \\
XGBoost                &  12.01 ± 0.28 &  1.21 ± 0.44 &  0.20 ± 0.02 &    \textbf{0.26 ± 0.16} &   1.63 ± 0.40 &   16.76 ± 6.68 &  14.99 ± 0.98 &  1.824 \\
KNN                    &  13.39 ± 0.37 &  1.17 ± 0.43 &  0.39 ± 0.03 &  68.46 ± 51.11 &   4.25 ± 0.96 &  60.76 ± 14.19 &  17.37 ± 1.26 &  1.752 \\
KNN-Scaled              &  12.54 ± 0.57 &  1.20 ± 0.44 &  0.19 ± 0.02 &    2.66 ± 1.92 &   4.14 ± 0.94 &   40.25 ± 7.00 &  15.41 ± 1.21 &  1.770 \\
LL                     &  14.07 ± 0.18 &  1.20 ± 0.46 &  0.33 ± 0.07 &  50.83 ± 14.81 &   7.04 ± 1.39 &  51.40 ± 10.50 &  16.69 ± 1.10 &  0.792 \\
LL-Scaled               &  12.90 ± 0.29 &  \textbf{1.10 ± 0.43} &  0.21 ± 0.02 &    2.47 ± 1.70 &   3.53 ± 1.24 &   32.54 ± 6.33 &  14.57 ± 0.96 &  0.786 \\
Tabnet               &  17.02 ± 2.68 &  1.22 ± 0.38 &  0.39 ± 0.04 &    2.83 ± 2.59 &   9.98 ± 3.04 &  43.73 ± 14.59 &  \textbf{12.97 ± 0.92} &    1.304 \\
\midrule
DNNR        &  12.31 ± 0.35 &  1.12 ± 0.47 &  \textbf{0.19 ± 0.02} &    1.05 ± 0.63 &   2.83 ± 0.60 &  36.52 ± 18.03 &  13.34 ± 0.96 &  \textbf{0.708} \\
DNNR-2 ord.  &  \textbf{11.64 ± 0.44} &  1.24 ± 0.50 &  0.22 ± 0.02 &    0.48 ± 0.43 &   2.30 ± 0.48 &  28.35 ± 13.47 &  15.61 ± 1.94 &  0.727 \\
\bottomrule
\end{tabular}
\end{table*}

\fakeparagraph{Benchmark Datasets: } 
The goal of this benchmark is to inspect DNNR's performance 
on eight real-world regression datasets:
Yacht, California, Protein, Airfoil, Concrete, Sarcos, CO2 Emissions, and NOX emissions.
The last two datasets are part of the Gas Emission dataset.
All datasets were taken from the UCI repository \cite{Du2019uci}, 
except California \cite{pace1997california} and Sarcos \cite{gpml}.
These datasets were also used in previous work \cite{dgp2016}.
Some of the datasets also come with discrete features (Yacht, Airfoil),
which challenges DNNR's assumption of continuity.
All the datasets were evaluated using a 10-fold split, 
except for the Sarcos dataset which comes with a predefined test set.
Additionally, we fixed the data leakage in the Sarcos dataset by removing all test data points from the training set.

In Table \ref{tab:results_datasets}, we report the averaged mean squared error (MSE) over the 10-folds for each dataset and model.
Overall, CatBoost is the best performing method.
We find that DNNR achieves the best performance on the 
Sarcos and California datasets and the second order achieves the best performance on Protein.
For NOxEmissions, CO$^2$Emissions, and Protein, DNNR is within 5\% percentage difference to the best performing method 
(see Table \ref{tab:percentage_difference}).
Discrete features violate DNNR's assumptions, which  
affect its performance, as the results on Airfoil and Yacht indicate. 

Discrete features can render the Taylor approximation meaningless, e.g. 
all neighbors may have the same value in one dimension rendering the gradient zero, or a linear approximation may not be sufficient when values exhibit large jumps. Interestingly, the second-order DNNR yields better results on the Airfoil and Concrete datasets, presumably because the second order is a better approximator of sharp gradients.

On the other datasets, DNNR delivers a significant improvement over KNN
and also over KNN with DNNR feature scaling (KNN-Scaled).

\begin{figure}[t]
    \centering
    \includegraphics{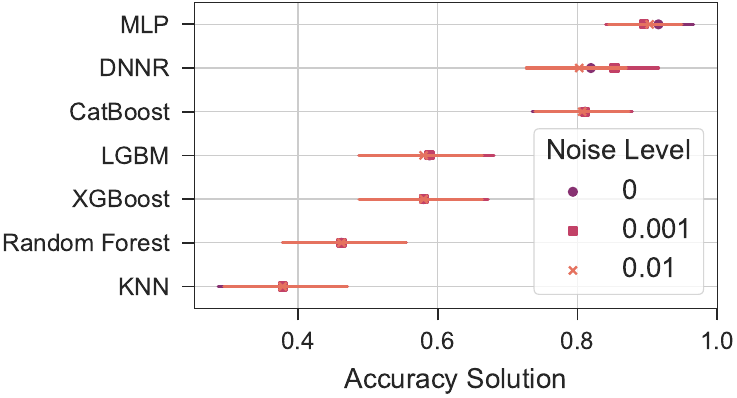}
    \caption{Accuracy on the Feynman Symbolic Regression Database under three levels of noise. The marks show the percentage of solutions with  $R^2 > 0.999$ .
    The bars denote $95\%$ confidence intervals.
    }
    \label{fig:feynman_accuracy}
\end{figure}
\begin{figure}[t]
    \centering
    \includegraphics{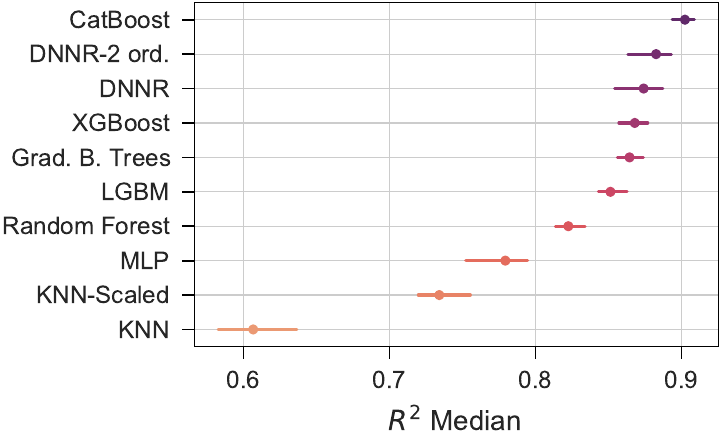}
    \caption{Results on the PMLB benchmark. The markers show the median $R^2$ performance over all datasets runs. Horizontal bars indicate the 95\% bootstrapped confidence interval.}
    \label{fig:pmlb_r2_median}
\end{figure}

\fakeparagraph{Feynman Benchmark}
As second benchmark, we selected the Feynman Symbolic Regression Database, which consists of 119 datasets sampled from classical and quantum physics equations 
\cite{Udrescu2020AIFA}. 
These equations are continuous differentiable functions. The 
difficulty can be increased by adding noise.

The evaluation for the Feynman benchmark was executed with 10 different splits for each dataset and noise level (std=0, 0.001, 0.01)
-- similar to \cite{cava2021contemporary}.
For the first split, we divided the data into 70/5/25\% train, validation, and test sets. 
The hyperparameter tuning was done with validation data of the first split.
Then the models were refit using the best parameters and evaluated on the 25\% test set.
Subsequent splits (75/25\% train/test) then used these hyperparameters.

For the Feynman benchmark, accuracy is defined as the percentage of 
 datasets which were solved with a coefficient of determination $R^2 > 0.999$. We report this accuracy in Figure \ref{fig:feynman_accuracy}.
DNNR is the second-best performing method after MLP. 
CatBoost's performance is also notable with an accuracy of more than 80\%.
The different noise levels had minor effects on the methods. 

\fakeparagraph{PMLB Benchmark}
The PMLB benchmark contains real and synthetic datasets with
categorical features, discrete targets, and noisy data in general.  
In total, we used 133 PMLB datasets.
The evaluation setup for the PMLB datasets was similar to the Feynman benchmark.
Figure \ref{fig:pmlb_r2_median} shows the median $R^2$ performance of the different models with a 95\% confidence interval. 
CatBoost is the best performing method with an $R^2$ median $> 0.9$ with DNNR second order closely in second position.
DNNR, XGBoost, and Gradient Boosted Trees perform similarly well.
The worst-performing method is KNN regression. While adding feature weighting (KNN-Scaled) improves the $R^2$ median considerably by over 0.1, only DNNR's additional use of gradient information yields results comparable to gradient boosting methods.

\subsection{Ablation 
}
\label{subsec:ablation}

In the previous evaluations, we already included KNN-Scaled to measure the effect of the scaling versus the gradient information.
We dissected DNNR even further and tested various design alternatives:
such as scaling of the neighborhood (MLKR \cite{weinberger2007metric}) and regularization on the gradient estimation (Lasso \cite{ausset2021nearest}).
We based this analysis on the Airfoil, Concrete, and 5000 samples from Friedman-1 datasets (see Section \ref{subsec:effect_samples_noise}). 
As before, we conducted a hyperparameters sweep for each model configuration and used a 10-fold validation for each dataset. 

For the Concrete and Friedman-1 
datasets, using only gradient information and no feature scaling (DNNR-Unsc.) already improved over KNN's performance. 
For the Airfoil dataset, which contains categorical features, using gradients without scaling (DNNR-Unsc.) lead to worse results. 
We find that MLKR improves KNN's performance better than DNNR's scaling for KNN. However, when using gradient estimation, MLKR is less suitable
as can be seen in the difference between DNNR and DNNR-MLKR on Airfoil and Concrete. Furthermore, 
we would like to note that MLKR is also computationally more expensive than DNNR's scaling as they use Gaussian kernels which results in a runtime quadratic in the number of samples.
These results highlights that while the gradient information might be helpful for unscaled neighborhoods, scaling yields better gradients and results in better approximations.

Using Lasso regularization on the gradient estimation as done in \cite{ausset2021nearest} did not perform well. 
We speculate that the regularization limits the gradient estimation.
Future work might test if DNNR benefits from Lasso regularization in high-dimensional problems as motivated by the authors.

\begin{table}
    \vspace{-0.5cm}
    \centering
    \caption{Ablation study. We report the MSE for different variations of DNNR for three datasets.
    \vspace{0.05cm}
    }
    \label{tab:ablation}
\footnotesize
\setlength{\tabcolsep}{0.47em} 
\begin{tabular}{llll}
\toprule
{} &         Airfoil &                 Concrete &    Friedman-1 \\
\midrule
DNNR            &     2.83 ± 0.60	 &            36.52 ± 18.03 &  \textbf{0.01 ± 0.00} \\
DNNR-Unsc. &     4.82 ± 0.74 &            49.97 ± 13.59 &  1.03 ± 0.15 \\
KNN-MLKR        &     3.32 ± 0.84 &             36.85 ± 9.89 &  0.40 ± 0.07 \\
KNN-Scaled      &     4.14 ± 0.94 &             40.25 ± 7.00 &  7.27 ± 0.53 \\
DNNR-MLKR       &     3.20 ± 1.08 &  1.5e13 ± 4e13 &  0.03 ± 0.00 \\
KNN             &     4.25 ± 0.96 &            60.76 ± 14.19 &  3.93 ± 0.43 \\
DNNR Rand.     &    24.38 ± 3.44 &           115.14 ± 20.52 &  6.07 ± 0.56 \\
DNNR-Lasso      &     5.25 ± 1.05 &             40.24 ± 8.43 &  7.33 ± 0.52 \\
DNNR-2 ord.   &     \textbf{2.30 ± 0.48} &             \textbf{28.35 ± 13.47} &  \textbf{0.01 ± 0.00} \\
\bottomrule
\end{tabular}
\end{table}

\subsection{Effect of noise, \#samples, and \#features}
\label{subsec:effect_samples_noise}

This analysis investigates how different data properties affect the model's performance.
Such an analysis requires a controlled environment:
we used the Friedman-1 dataset \cite{friedman1991multi}. This dataset is based on the following equation:
\begin{equation}
    \label{eq:friedman1} 
    \begin{split}
        y(x) = 10 x_{3} + 5 x_{4} & + 20 \left(x_{2}\!-\! 0.5\right)^{2} \\
                & + 10 \sin{\left(\pi x_{0} x_{1} \right)}
                  + s \epsilon,
    \end{split}
\end{equation}
where $x_i$ is uniformly distributed in [0, 1], the noise $\epsilon$ is sampled from a standard normal distribution, and $s$ controls the variance of the noise. Unimportant features can be added by simply sampling $x_j \sim U(0, 1)$. 
Friedman-1 allowed us to test the models under different sampling conditions: 
the number of samples,
the magnitude of noise, 
and the number of unimportant features. 

As defaults, we choose the number of samples = 5000,
the number of features = 10, and the noise level = 0.
Besides DNNR, we also evaluated Gradient Boosted Trees, CatBoost, Random Forest, MLP, and KNN.
For each setting, we run a 5-fold evaluation 
(the hyperparameters of each methods are fitted on
the first fold and then fixed for the remaining 4).

We report the effect of each condition in Appendix Figure \ref{fig:effect_of_data_properties}.
For the number of samples, we note that DNNR's error declined rapidly. Second-best is CatBoost.
For noise, we observed two groupings. While one group (MLP \& KNN) performed poorly, their MSE did not increase when adding noise. For the better performing group (DNNR, Gradient Boosted Trees, CatBoost, Random Forest), the error did increase when adding noise. In this group, DNNR performed the best for low noise levels, but 
was beaten by CatBoost slightly for higher levels of noise.

Increasing the number of unimportant features impacted KNN and MLP particularly. DNNR dropped from being the best method to sharing the second place with Gradient Boosted Trees as the feature scaling cannot entirely mitigate the effect of unimportant features. Tree-based methods were barely affected, as they are adept at handling unimportant features and operate on the information gain of each feature. 

\subsection{Application of the theoretical bound}
\label{subsec:empirical_validation_theory}

\begin{figure}[t]
    \centering
    \includegraphics{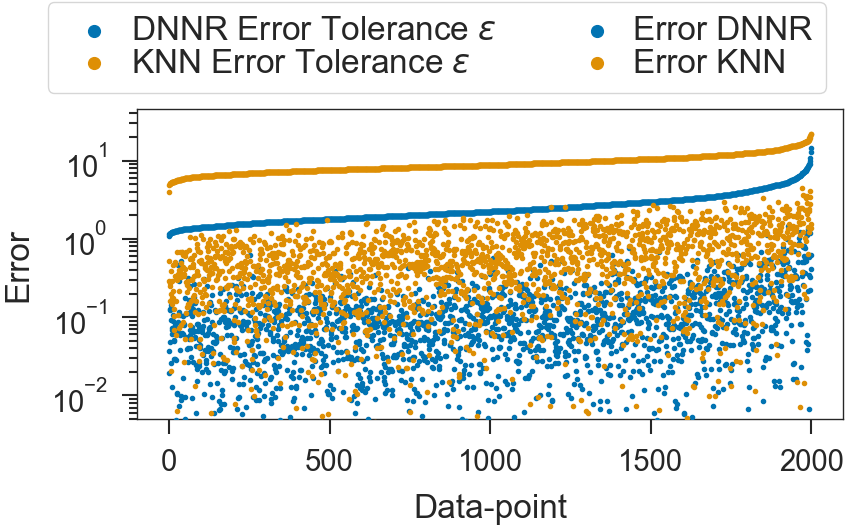}
    \caption{
        Comparison between the error bound of KNN (yellow) and DNNR (blue). 
        On the x-axis, all test data points sorted by their error bounds are plotted.
        The DNNR performs better than KNN and the DNNR error bound is also lower.
    }
    \label{fig:bounds_main}
\end{figure}
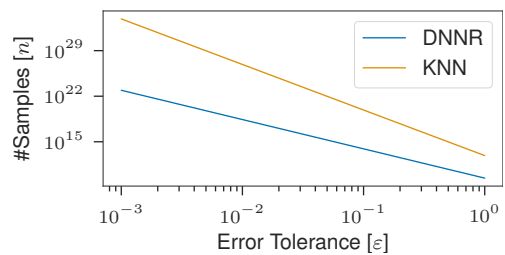
\begin{figure}[t]
    \centering
    \input{figures/bounds/error_tolerance.pgf}
    \caption{
        Depicts the error tolerance versus the number of training samples for the Friedman-1 dataset. KNN requires multiple orders more training points to guarantee the same error tolerance. 
    }
    \label{fig:error_tolerance_vs_n_samples}
    \vspace{-0.3cm}
\end{figure}

In this evaluation, we analyze the error bounds from section \ref{sec:theory}. As dataset, we use the Friedman-1 dataset introduced before in section \ref{subsec:effect_samples_noise}. 
The synthetic dataset allows sampling arbitrary points, thereby we can simulate also very dense neighborhoods.

First, we apply Theorem \ref{theorem:DNNR_pointwise_error} to the dataset.
We pick a probability tolerance of 0.95 ($\delta=0.05$), and a Lipschitz constant for equation \eqref{eq:friedman1} of $\vartheta_{\max}=40$. 
For DNNR, we estimate a value of $\tau \approx 5.59$ from the data.
We show the dependency between error tolerance the number of samples required in Figure 
\ref{fig:error_tolerance_vs_n_samples}. 
In this exemplary calculation, KNN requires multiple orders 
of magnitude more training data than DNNR to achieve the same theoretical guarantee on 
the error tolerance.
Still, even DNNR  would require an unrealistic amount of samples, e.g. around $10^{15}$ for an error tolerance of $\varepsilon=0.1$.

As a more practical application, we investigated how 
local conditions, e.g. distance to the neighbors and the Lipschitz constant,
influence the local prediction error.
Therefore, we sampled a realistically sized train and test set (10.000 and 2.000 samples) and then compared the error tolerance for KNN and DNNR
according to equations \eqref{eq:dnnr_error_tolerance} and \eqref{eq:knn_error_tolerance}.
For both methods, we choose $k=7$, and for the number of neighbors to approximate the gradient, we used $k'=32$. 
We purposely violate the conditions on $k$ and $n$ of Theorem \ref{theorem:DNNR_pointwise_error}, as we want to analyze how applicable the estimated error tolerances are in a more realistic setting.
In Figure \ref{fig:bounds_main}, we sorted the 2000 test points according to the error tolerances of KNN and DNNR respectively. 
For both methods, we see that the error tolerances were strictly bound 
the actual errors by a gap of around one order of magnitude. We also observe that the 
error tolerances were correlated with  the actual errors, e.g. as the error tolerances decrease, so does the actual error.

\subsection{Feature importance \& Inspecting a prediction}

\begin{figure}[t]
    \begin{subfigure}[t]{0.41\linewidth}
        \vskip 0pt

        \includegraphics[scale=0.94]{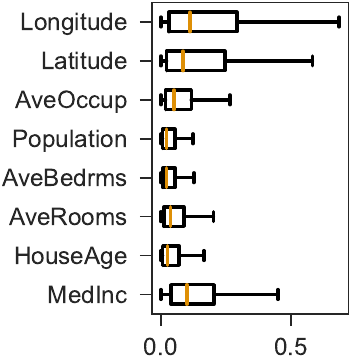}
        \caption{Feature relevances}
        \label{fig:local_feature_importance}
    \end{subfigure}
    \begin{subfigure}[t]{0.58\linewidth}
        \vskip 0pt
        \includegraphics[scale=0.94]{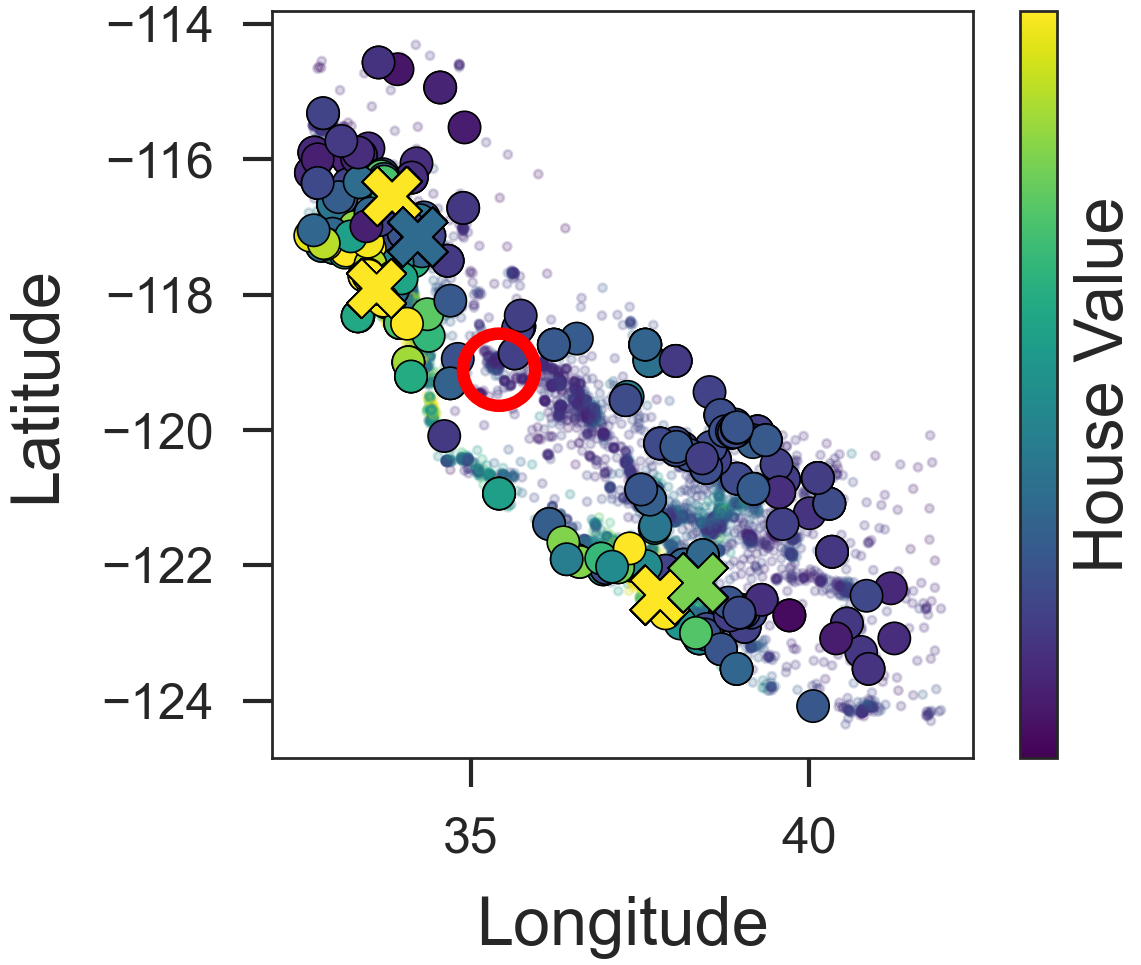}
        \caption{Failure (error of 4.85)}
        \label{fig:inspection_of_neighbors}
    \end{subfigure}
    \caption{
        \textbf{(a)} Feature relevance of DNNR Unscaled on the California Housing dataset. 
        \textbf{(b)} A failure of DNNR Unscaled.
        The red circle marks the query point. The prediction is done by approximating the gradient locally for each nearest neighbors (crosses). The circles visualize the points used for gradient approximation.
        As DNNR Unscaled weights each dimension equally,
        it does not use spatially nearby points, 
        even though longitude and latitude are scored important. 
    }
\end{figure}

DNNR allows inspecting which neighbors were used for the prediction 
and how they contributed.
For the following exemplary inspection, we use the California Housing dataset \citep{pace1997california}.
The dataset's dependent variable is the median house value per block group (a block group is an aggregation of a local area). 
The eight observational variables represent the location, 
median income, and information about the houses, such as average rooms or occupation.  
For this study, we analyzed DNNR Unscaled, i.e. DNNR without the feature scaling. We omitted the feature scaling, as the feature relevance would be impacted by the scaling and the DNNR with scaling is also performing so well that the error would be minor to inspect.

First, we can provide a simple local feature relevance score by multiplying the 
estimated gradient with the difference in the input: 
\begin{equation}
    \bm{\xi}_m = |(\vx - \vx_m) \odot \hat \gamma_m|,
\end{equation}
where $x_m$ is the point where we estimate the gradient and $\hat \gamma_m$ the locally fitted gradient. 
This formulation of feature importance is analogous to a linear model where one would take $w \odot x$ 
\citep[sec. 5.1.]{molnar2019}.
It is a known property that the gradient reflects the model's sensitivity and can be used for feature importance, and variable selection \cite{mukherjee2006covariance,guyon2003introduction}.
We show the distributions of the local feature importance in Figure \ref{fig:local_feature_importance}. 

The most important dimensions are longitude, latitude, and median income. 
We validate the local feature importance by applying it to variable selection. 
Using all dimensions, we get an MSE of around 0.34 (this is lower than
in Table \ref{tab:results_datasets} as we do not use feature weighting).
When deleting the three most important dimensions, the MSE increases to 0.99.
However, keeping only the most important dimensions, the MSE slightly improves to 0.33.
Therefore, we conclude that the feature importance has found the most important dimensions.

We now move on to inspect a failure case of DNNR.
In Figure \ref{fig:inspection_of_neighbors}, we show how the neighborhood of a badly predicted point can be inspected. The prediction (red circle) is off by 4.85. 
From looking at the projection of the data to the longitude and latitude dimensions,
we can see that the prediction is based on points (crosses) far away from the query. 
These points might have a similar number of bedrooms but differ in the location.
As we found in the previous experiment, the latitude/longitude belong to the most important features.
This inspection motivates the feature scaling once again, as DNNR Unscale weights all dimension equally, it selected the nearest neighbors using  less relevant dimensions.

\section{Conclusion, Limitations, and Future Work}

\fakeparagraph{Conclusion}
DNNR showed that local datapoint gradients carry valuable information for prediction and can be exploited using a simple Taylor expansion to provide a significant performance boost over KNN regression. In large-scale evaluations, DNNR achieved comparable results to state-of-the-art complex gradient boosting models. An advantage of DNNR's simplicity is that we can obtain error bounds by extending KNN's theory. Our theoretical analysis illustrates the benefits of using DNNR over KNN.
DNNR strikes a good balance between performance and transparency and may therefore be the method of choice in problems with elevated requirements for the system's interpretability.

\fakeparagraph{Limitations}
Our evaluation of DNNR points to a potential limitation on discrete data. When features or targets have the same value, the gradient is zero and DNNR falls back to KNN's decision. 
On partially discrete datasets DNNR performs always at least as well as KNN, see e.g. the results of DNNR and KNN on the Yacht dataset (Table \ref{tab:results_datasets}). 
DNNR also inherits some limitations from KNN, such that the L2-metric might not represent similarities optimally.

\fakeparagraph{Future Work}
DNNR was designed for regression tasks but could be adapted for classification as well, e.g. by fitting the gradients of a logistic function as in \cite{coordclassif2006} or via label smoothing \cite{muller2019labelsmoothing}.
An interesting future direction may be in extending DNNR specifically to symbolic regression by utilizing the estimated gradient information. 
Future work could also explore the use of DNNR for data augmentation, where points could be sampled, and the label computed estimated with the local gradient.
Another research direction would be to tighten the theoretical bounds or to study the effect of scaling from a theoretical perspective.

\subsection*{Acknowledgments}

We thank Maximilian Granz, Oana-Iuliana Popescu, Benjamin Wild, Luis Herrmann, and David Dormagen for fruitful discussion and feedback on our manuscript. 
We are also grateful for our reviewers' feedback. 
LS was supported by the Elsa-Neumann Scholarship of the state of
Berlin.  We thank the  HPC Service of ZEDAT, Freie Universit\"at Berlin, for generous allocations of computation
time \cite{Bennett2020}.

\bibliography{references}
\bibliographystyle{icml2022}

\appendix

\onecolumn

\section{Figures}

\begin{figure}[H]
    \begin{subfigure}[t]{0.23\linewidth}
    \vskip 0pt
        \includegraphics[scale=0.80]{figures/inspection/california_neighbors_bad_1305.png}
        \caption{Sample 1305: Error of 4.85}
    \end{subfigure}
    \begin{subfigure}[t]{0.26\linewidth}
    \vskip 0pt
        \includegraphics[scale=0.80]{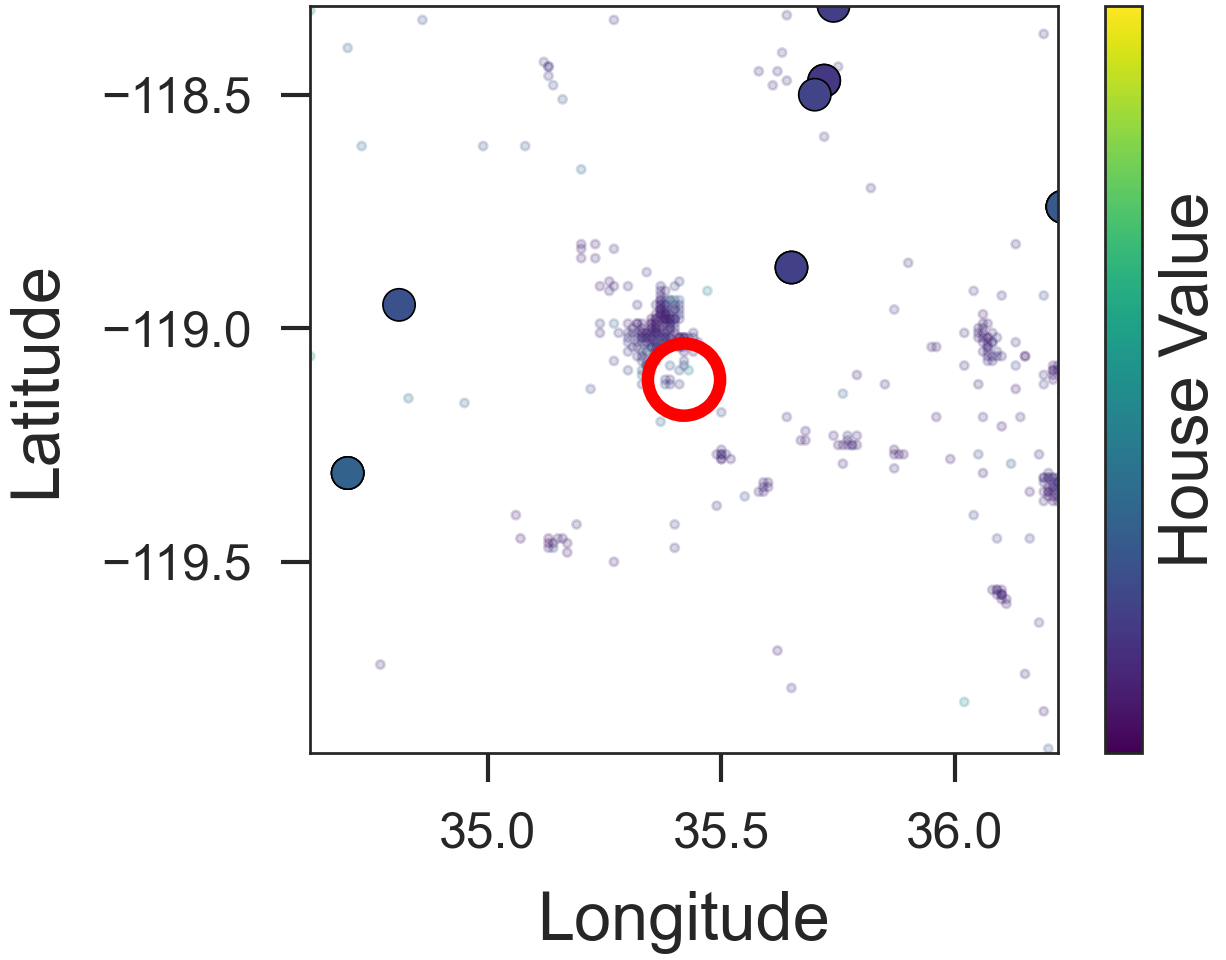}
        \caption{Zoom of 1305}
    \end{subfigure}
    \begin{subfigure}[t]{0.23\linewidth}
    \vskip 0pt
        \includegraphics[scale=0.80]{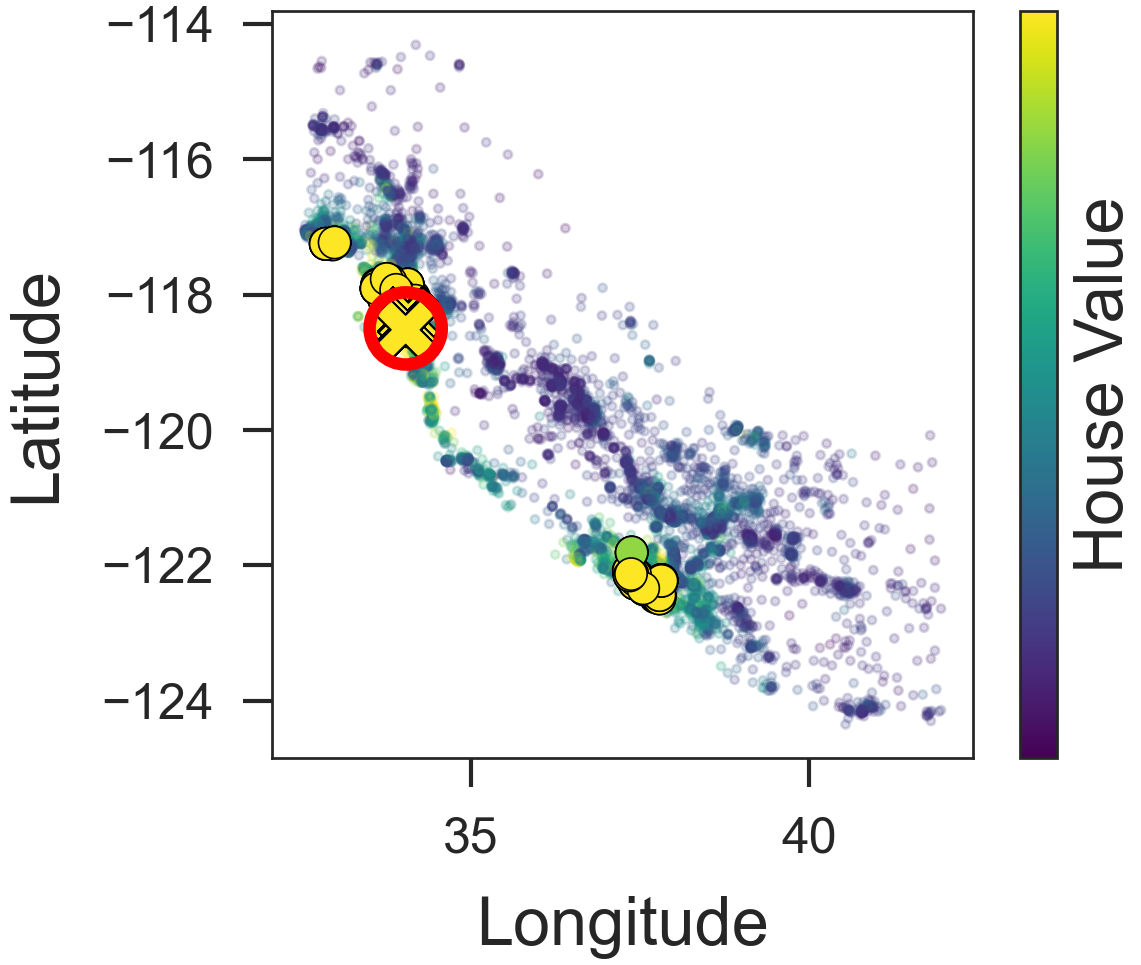}
        \caption{Sample 1220 (Error 0.02)}
    \end{subfigure}
    \begin{subfigure}[t]{0.26\linewidth}
    \vskip 0pt
        \includegraphics[scale=0.80]{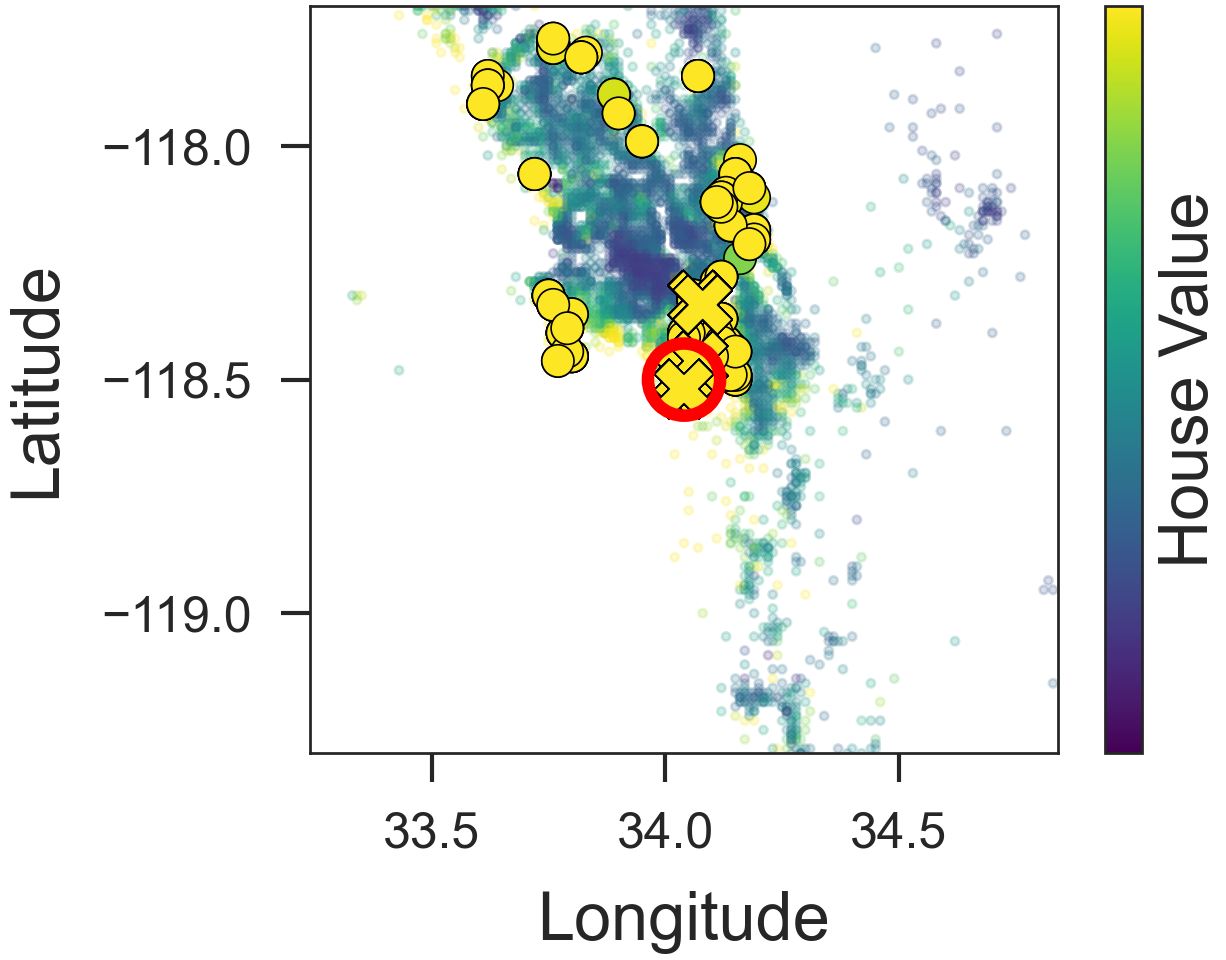}
        \caption{Zoom of 1220}
    \end{subfigure}
    
    \begin{subfigure}[t]{0.23\linewidth}
    \vskip 0pt
        \includegraphics[scale=0.80]{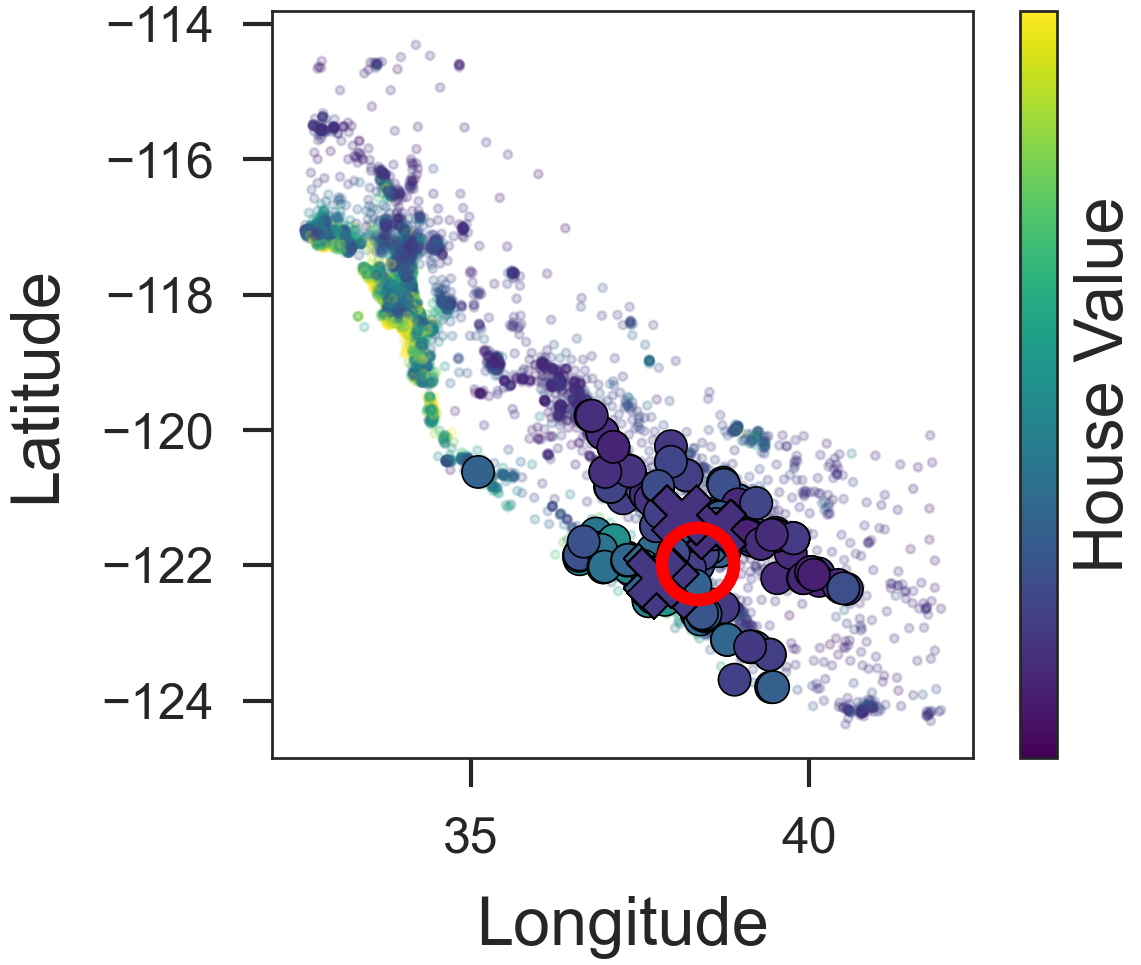}
        \caption{Sample 1081 (Error 0.13)}
    \end{subfigure}
    \begin{subfigure}[t]{0.26\linewidth}
    \vskip 0pt
        \includegraphics[scale=0.80]{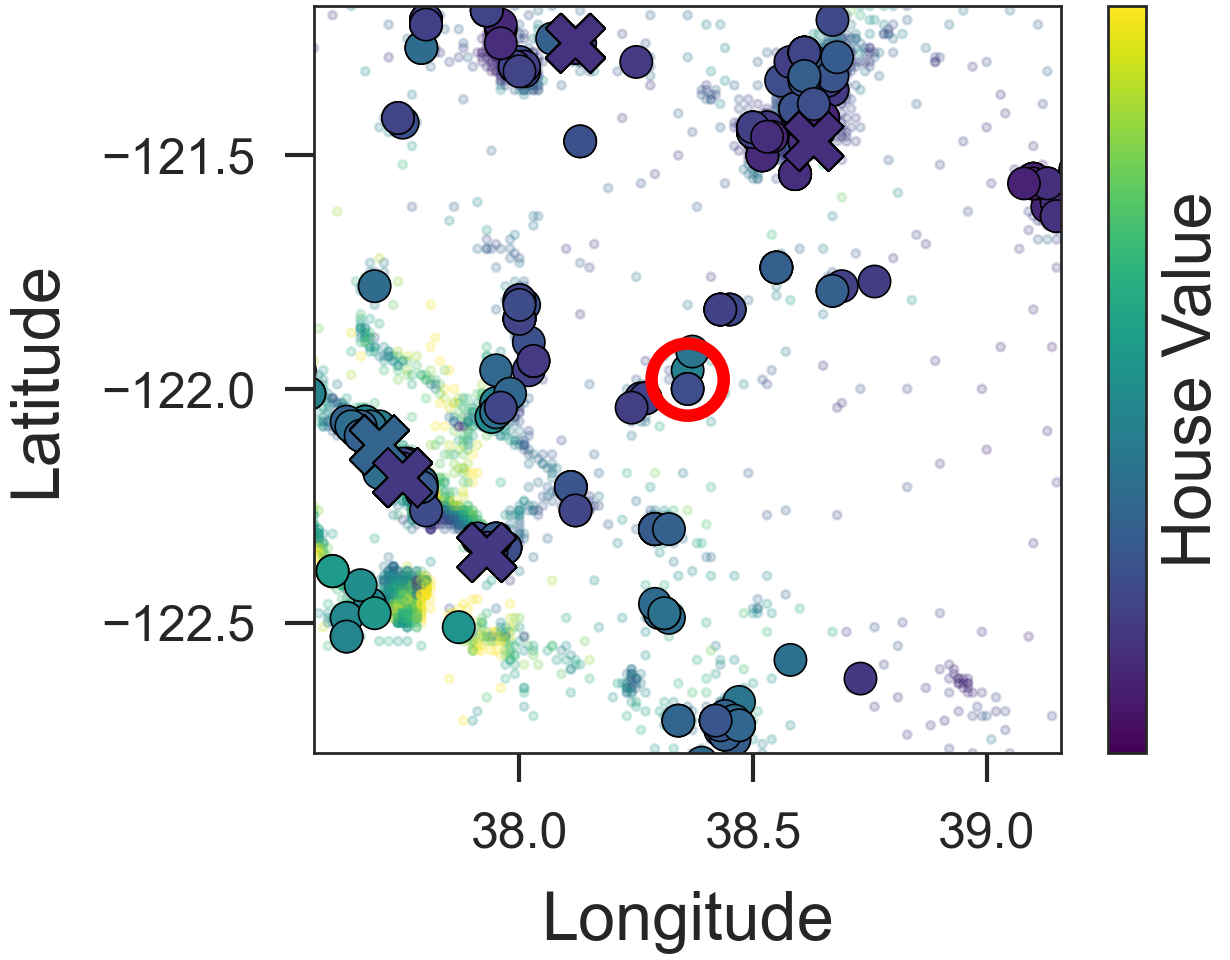}
        \caption{Zoom of 1081}
    \end{subfigure}
    \begin{subfigure}[t]{0.23\linewidth}
    \vskip 0pt
        \includegraphics[scale=0.80]{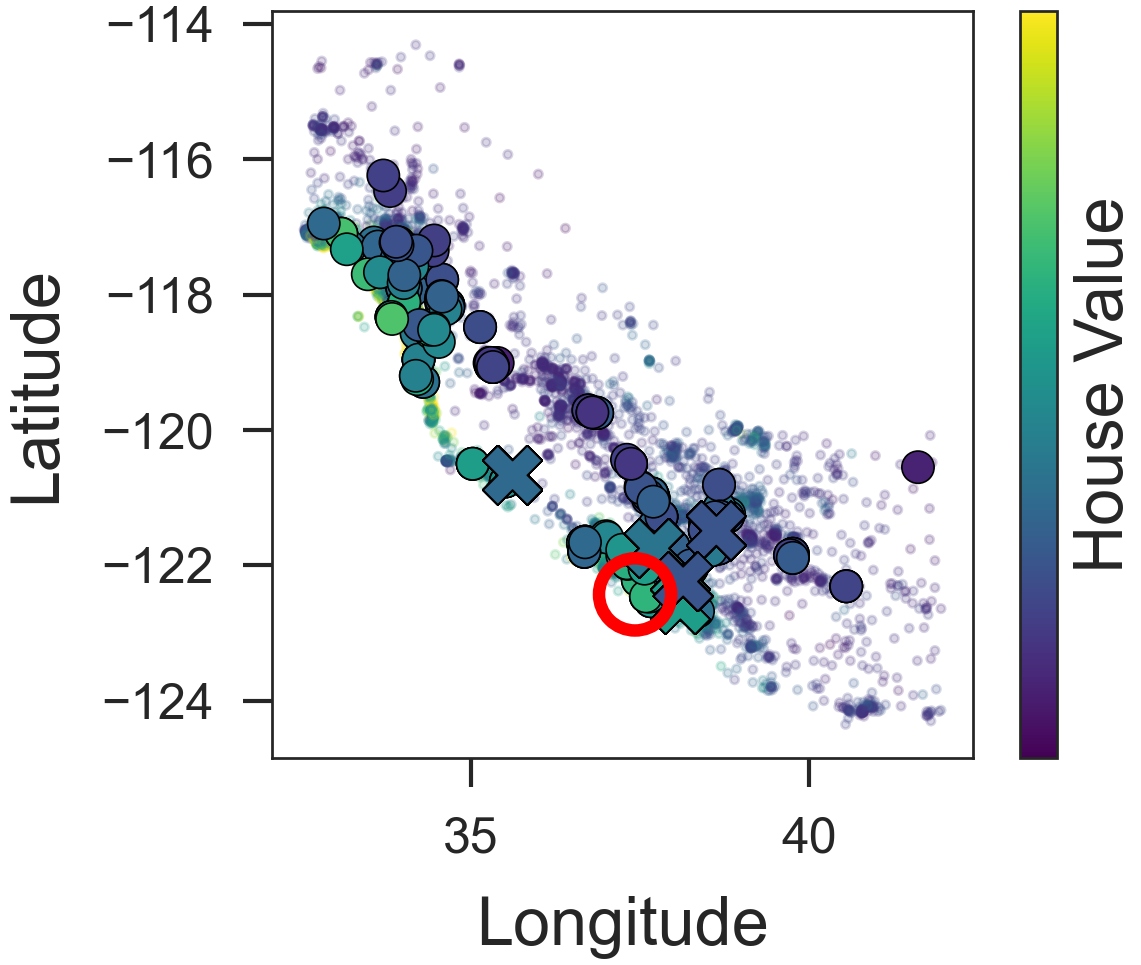}
        \caption{Sample 393 (Error 1.01)}
    \end{subfigure}
    \begin{subfigure}[t]{0.26\linewidth}
    \vskip 0pt
        \includegraphics[scale=0.80]{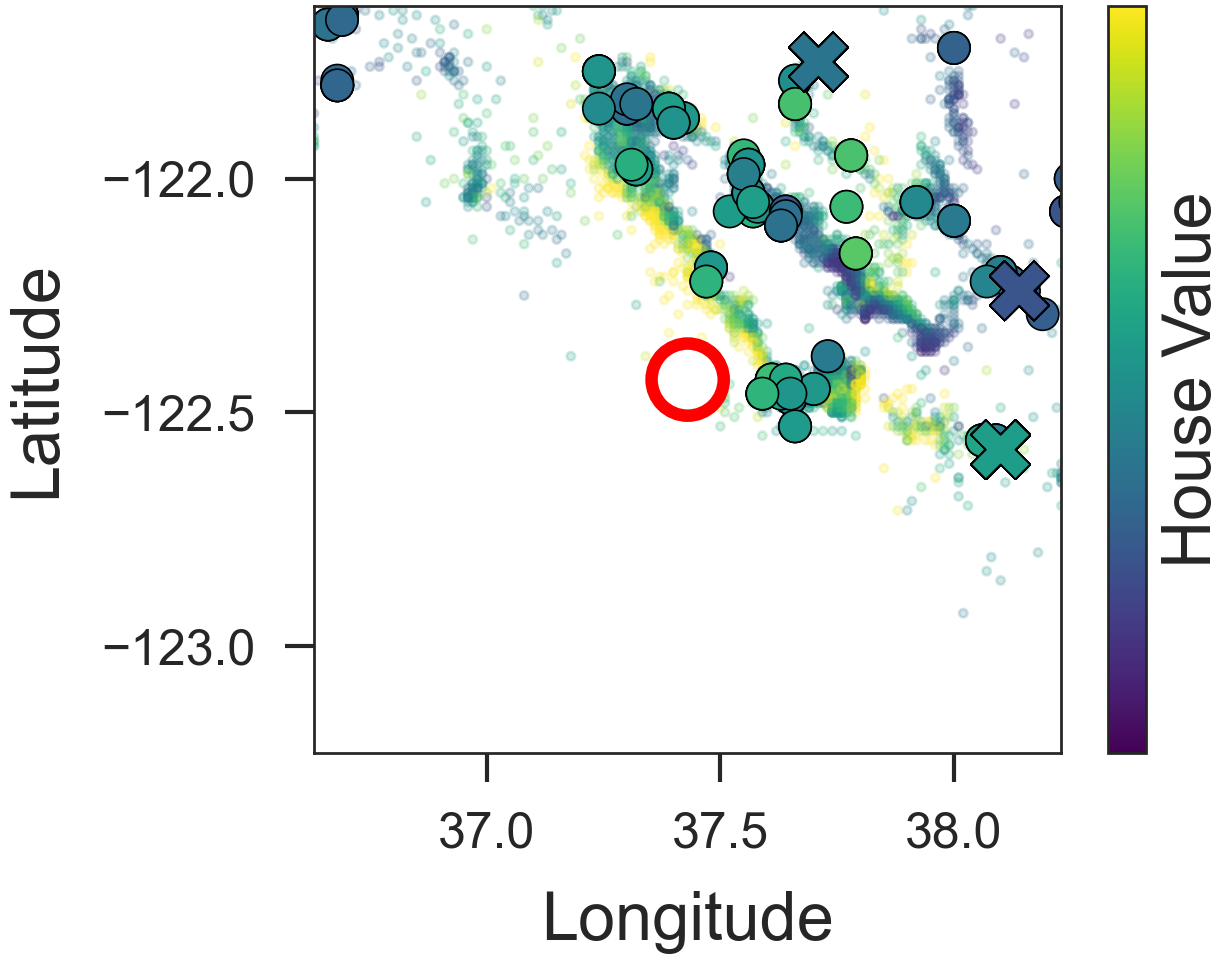}
        \caption{Zoom of 393}
    \end{subfigure}
    \caption{
        The selected neighbors for different queries on the California dataset.
        The red circle marks the query point. 
        The prediction is done by approximating the gradient locally for each nearest neighbors (crosses). The filled circles visualize the points used for gradient approximation.
        The neighbors are further away for higher errors \textbf{(a) \& (g)} while close to the query for lower errors \textbf{(c) \& (e)}.
    }
\end{figure}

\begin{figure}[H]
    \centering
    \begin{subfigure}[b]{0.45\textwidth}
         \centering
         \includegraphics{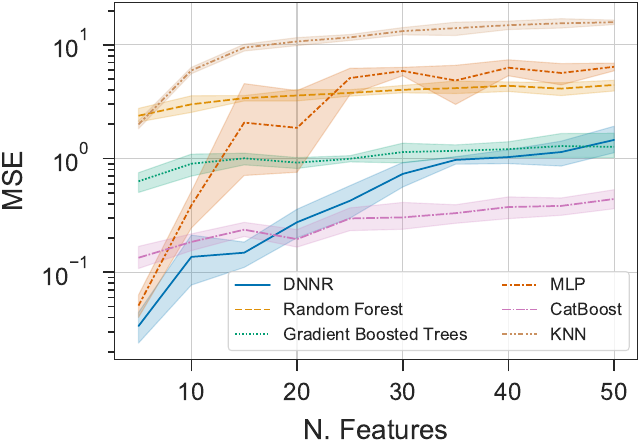}
        \caption{ Number of Features}
        \label{fig:freidman_n_features}
    \end{subfigure}
    \begin{subfigure}[b]{0.45\textwidth}
         \centering
        \includegraphics{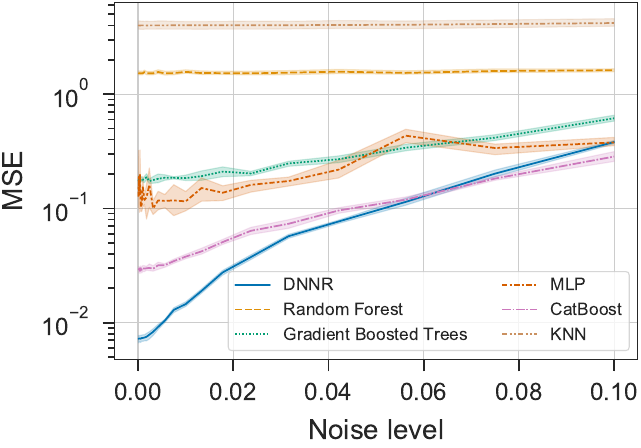}
        \caption{Noise Level}
        \label{fig:freidman_noise_level}
    \end{subfigure}
    
    \begin{subfigure}[b]{0.45\textwidth}
         \centering
        \includegraphics{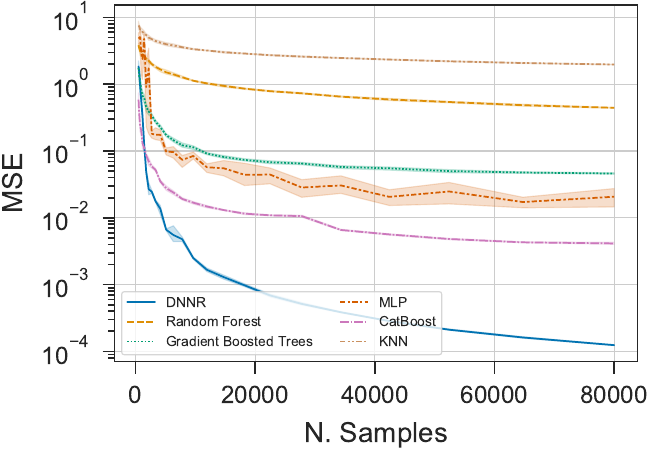}
        \caption{Number of Samples}
        \label{fig:freidman_n_samples}
    \end{subfigure}
    
    \caption{
        The effect of different parameters of the dataset. The results are obtains on the Friedman-1 dataset.
        The confidence intervals denote the standard derivation over multiple folds.
    }
    \label{fig:effect_of_data_properties}
\end{figure}

\newpage


\section{Tables}
\begin{table}[H]
\centering
\caption{Mean percentage difference from the best performing model on each dataset. }
\setlength{\tabcolsep}{0.47em} 
\label{tab:percentage_difference}
\begin{tabular}{lllllllll}
\toprule
{} & Protein & CO$^2$Emission & California &  Yacht & Airfoil & Concrete & NOxEmission& Sarcos\\

\midrule
CatBoost          &       1.54\% &           1.09\% &          0.00\% &      0.00\% &       0.00\% &        0.00\% &          12.91\% &     85.45\% \\
Grad. B. Trees    &       4.42\% &          13.71\% &          7.94\% &     32.14\% &      39.32\% &       14.53\% &          20.26\% &    156.07\% \\
KNN               &      15.03\% &           6.63\% &        108.47\% &  27065.48\% &     236.23\% &      334.12\% &          33.91\% &    147.46\% \\
KNN-neigh         &       7.76\% &           8.90\% &          2.65\% &    953.97\% &     227.69\% &      187.56\% &          18.79\% &    150.00\% \\
LGBM              &       9.51\% &           6.09\% &          3.17\% &   2623.41\% &      60.36\% &       15.05\% &          20.74\% &    127.82\% \\
LL                &      20.93\% &           8.99\% &         72.49\% &  20072.22\% &     457.28\% &      267.28\% &          28.69\% &     11.86\% \\
LL-neigh          &      10.83\% &           0.00\% &         12.70\% &    880.95\% &     179.35\% &      132.46\% &          12.35\% &     11.02\% \\
MLP               &      23.22\% &          11.99\% &         84.13\% &   1623.02\% &     704.43\% &      151.99\% &          53.46\% &     80.37\% \\
Random Forest     &       1.39\% &           1.54\% &         21.69\% &    348.81\% &     122.15\% &       62.53\% &          26.08\% &    219.77\% \\
Tabnet            &      46.25\% &          10.63\% &        108.47\% &   1024.21\% &     689.79\% &      212.47\% &           0.00\% &     84.18\% \\
XGBoost           &       3.17\% &           9.99\% &          6.35\% &      2.78\% &      28.72\% &       19.73\% &          15.54\% &    157.63\% \\
\midrule
DNNR-2 ord. &       0.00\% &          12.35\% &         13.76\% &     88.89\% &      81.96\% &      102.52\% &          20.37\% &      2.68\% \\
DNNR   &       5.79\% &           1.27\% &          0.00\% &    315.08\% &     123.58\% &      160.93\% &           2.86\% &      0.00\% \\
\bottomrule
\end{tabular}

\end{table}

\section{Approximation Bounds}

\begin{table}[H]
    \centering
    \begin{tabular}{l l}
        \toprule
        Notation &  Meaning \\
        \midrule
        $x$    
        & An input vector 
        \\
        $X$    
        & The random variable of the input 
        \\
        $Y$    
        & The random variable of the output 
        \\
        $k$    
        & The number of nearest neighbors 
        \\ 
        $k'$   
        & The number of neighbors to estimate the gradient 
        \\ 
        $\eta(x)$ 
        & The expected target value: 
        $\eta(x) = \E[Y|X=x]$  
        \\ 
        $\nabla \eta(x)$ 
        & The gradient of $\eta(x)$ w.r.t. $x$
        \\ 
        $\eta_{\text{KNN}}(x)$ 
        & Estimated regression function for KNN regression
        \\
        $\eta_{\text{DNNR}}(x)$ 
        & Estimated regression function for DNNR
        \\
        $\text{nn}(x, k)$ 
        & Returns the indices of the $k$ nearest neighbors of $x$
        \\
        $\hat Y_{m, x}$ 
        & Estimated regression value for $x$ from point $X_m$  
        \\
        $\vartheta_{\max}$ 
        & Maximum Lipschitz constant 
        \\
        $A$    
        & Matrix to estimate the gradient using OLS
        \\
        $\sigma_1$  
        & Smallest singular value of $A$ 
        \\
        $C^\mu$ 
        & Set of $\mu$ times differentiable continuous functions
        \\
        $(v \cdot \nabla)^\mu f$ 
        &  The $\mu$-order directional derivative of f w.r.t $v$
        \\
        $\E_n[.]$ & Expectation over $n$ training points  $(X_i, Y_i)_{
        1 \le i \le n}$
        \\
        $ {\hat \gamma}_m  $ 
        & Locally estimated gradient for point $m$.
        \\
        $ {\hat \omega}_m  $ 
        & Locally estimated gradient and higher order terms
        \\
        $ \varepsilon  $ 
        & The error tolerance
        \\
        $ \delta  $ 
        & Probability tolerance
        \\
        $ h_m  $ 
        & Distance between $x$ and point $X_m$ 
        \\
        $\rho(x, x')$ & Distance metric
        \\
        \bottomrule
    \end{tabular}
    \caption{Notation used in this work.}
    \label{tab:notation}
\end{table}

The proof of the approximation bounds of DANN extends the
the proof of KNN approximation bounds given in \citep[p.~68ff.]{Chen2018ExplainingTS}
by a Taylor Approximation. For the approximation of the gradient, we will
rely on results given in \citep{Turner2010ErrorBF}.

\fakeparagraph{ Notation} Our notation follows the one given in \citep{Chen2018ExplainingTS}:

\subsection{General Properties}

We list here some inequalities used in the later proof.

\fakeparagraph{Jensen's Inequality}
Given a random variable $X$ and a convex function $f(X)$ then:
\begin{equation}
    f(\E[X]) \le \E[f(X)].
\end{equation}

\fakeparagraph{Hoeffding's Inequality}
Let $X_1, \ldots, X_k$ be independent random variables bounded between $[a, b]$.
The empirical mean is given by: $\bar X = \frac{1}{k} (X_1 + \ldots + X_k)$.
Then:
\begin{equation}
    \P(\left| 
        \bar X - \E[\bar X] 
    \right| > t)
    \le  2\exp \left(
        -\frac
            {2kt^2}
            {(b - a)^2} 
    \right)
\end{equation}

\fakeparagraph{Chernoff Bound for Binomial distribution}
Let $X = \sum_{i=1}^n X_i$ be a sum of $n$ independent binary random variable each $X_i = 1$ with probability $p_i$.
Let $\mu = \E[X] = \sum_{i=1}^k p_i = n \bar p$, where $\bar p = \frac{1}{n}\sum_{i=1}^k p_i $. Then
\begin{equation}
    \P(X \le (1 - \delta) \mu) \le \exp(-\mu \delta^2 / 2).
\end{equation}

\subsection{Gradient Approximation}
\label{appendix:local_grad_approx}

Here, we first proof two lemmas for bounding the gradient.
They generalize the proof in \citep[section 3.1]{Turner2010ErrorBF} 
from 2 to $d$-dimensions.

The first Lemma bounds terms of a Taylor series and is need for the proof of Lemma \label{lemma:grad-estimation}.

\begin{lemma}
\label{lemma:least-sqaures-estimates}
Let $f: D \subset \R^d \to \R $ be of class $C^\mu$,  $a \in D$, and $\B(a) \subset D$
some neighborhood of a. Suppose that for $i = 0, 1, \ldots, n$ we have
$\frac{\partial^\mu f}{\partial^\mu a}  \in Lip_{\vartheta_i}(\B(a))$ and
$\vartheta_{\max} = \max_{i \in 1, \ldots \mu} \vartheta_i$.
Then for any $a + h \nu \in \B(a)$ with $||\nu|| = 1$
\begin{equation}
    \left| 
        \sum^\mu_{k=1} \frac{h^{k-1}}{k!} (\nu \cdot \nabla)^k f(a) 
        - \frac{f(a + h \nu) - f(a)}{h} 
    \right|
    \le 
    \frac{h^\mu}{(\mu+1)!} \vartheta_{\max}  ||v||_1^\mu.
\end{equation}
\end{lemma}

\textit{Proof  Lemma \ref{lemma:least-sqaures-estimates}.}
The proof starts with rearranging the Taylor Series which is
given here:
\begin{equation}
    f(a + h\nu) = f(a) 
    + h \frac{(\nu \cdot \nabla) f(a)}{1!}
    + \ldots
    + h^\mu \frac{(\nu \cdot \nabla)^\mu f(a)}{\mu!}
    + R_\mu.
\end{equation}
The remainder $R_\mu$ has the following form:
\begin{equation}
    R_\mu = \frac{h^{\mu+1}}{\mu!} \int_0^1 (1 - t)^\mu (\nu \cdot \nabla)^{\mu+1} f(a + th\nu) dt.
\end{equation}
By dividing by $h$ and rearanging some terms, we have:
\begin{equation}
\frac{f(a + h\nu) - f(a)}{h}
    - \sum_{k=1}^{\mu-1} \frac{h^{k-1}}{k!} (\nu \cdot \nabla)^k f(a) = 
    \frac{R_{\mu-1}}{h}.
\end{equation}
Now, we add the same quantity to both sides of the previous equation,
using $ \int_0^1 (1 - t)^{\mu-1} dt = \frac{1}{\mu}$:
\begin{align*}
    \frac{h^{\mu-1}}{\mu!} & (\nu \cdot \nabla)^\mu f(a)  
     - \frac{f(a + h\nu) - f(a)}{h}
     + \sum_{k=1}^{\mu-1} \frac{h^{k-1}}{k!} (\nu \cdot \nabla)^k f(a) \\ 
    & = \frac{h^{\mu-1}}{(\mu-1)!} \int_0^1 (1 - t)^{\mu - 1}
    \Big\{ 
        (\nu \cdot \nabla)^{\mu} f(a) 
        - (\nu \cdot \nabla)^{\mu} f(a + th\nu)
    \Big\} dt.
\end{align*}
Therefore, we have:
\begin{align*}
    & \left|
     \sum_{k=1}^{\mu-1} \frac{h^{k-1}}{k!} (\nu \cdot \nabla)^k f(a) 
     - \frac{f(a + h\nu) - f(a)}{h}
    \right|
    \\
    =    
    & \left| 
        \frac{h^{\mu-1}}{(\mu-1)!} \int_0^1 (1 - t)^{\mu - 1}
        \Big\{ 
            (\nu \cdot \nabla)^{\mu} (f(a) - f(a + th\nu))
        \Big\} dt
    \right|
    \\
    & = 
    \left| 
    \frac{h^{\mu-1}}{(\mu-1)!} \int_0^1 (1 - t)^{\mu - 1}
    \Big\{ 
        (\nu \cdot \nabla)^{\mu} (f(a) - f(a + th\nu))
    \Big\} dt 
    \right| \\
    & \le  
    \frac{h^{\mu-1}}{(\mu-1)!} \int_0^1 (1 - t)^{\mu - 1}
    \Big| 
        \left(
            \nu \cdot \nabla
        \right)^{\mu} \left(
            f(a) - f(a + th\nu)
        \right)
    \Big| dt 
     \\
    & \le  
    \frac{h^{\mu+1}}{(\mu+1)!}  \vartheta_{\max} \left| \nu \right|_1^\mu.
\end{align*}
In last step, we rewrote the $\mu$-th directional derivative 
using partial derivative and bounded it using the Lipschitz-continuity as follows:
\begin{align*}
    & |(\nu \cdot \nabla)^\mu (f(a) - f(a + th\nu))|  \\
    &= 
    \left|
    \sum_{k_1 + \ldots + k_d = \mu} \binom{\mu}{k_1,\ldots,k_d} 
        \left(
            \prod_{i=1}^{d} \nu_i^{k_i} 
        \right)
        \frac{\partial^\mu (f(a) - f(a + th\nu))}
            {\partial a_1^{k_1} \cdot \ldots \cdot \partial a_d^{k_d} }  
    \right| \\
    &= 
    \sum_{k_1 + \ldots + k_d = \mu} \binom{\mu}{k_1,\ldots,k_d} 
        \left(
            \prod_{i=1}^{d} 
            \left|
                    \nu_i^{k_i} 
            \right|
        \right)
        \frac{
            \partial^\mu 
                \left|
                    (f(a) - f(a + th\nu))
                \right|
            }
            {\partial a_1^{k_1} \cdot \ldots \cdot \partial a_d^{k_d} } 
    \\
    & \le 
    \sum_{k_1 + \ldots + k_d = \mu} \binom{\mu}{k_1,\ldots,k_d} 
        \left(
            \prod_{i=1}^{d} 
            \left|
                    \nu_i^{k_i} 
            \right|
        \right)
        \vartheta_{\max} | a - a + th\nu|
    \quad \text{(Lipschitz continuity of order $\mu$)}
    \\
    & =
    \vartheta_{\max} |th\nu|
    \left(
        \left| \nu_1 \right|
        + \ldots +
        \left| \nu_d \right|
    \right)^\mu
    \quad \quad 
    \text{(Multinomial theorem)}
    \\
    & =
    \vartheta_{\max} |th|
        \left| \nu \right|_1^\mu.
    \quad 
    \quad 
    \\
\end{align*}
This finishes the proof of Lemma \ref{lemma:least-sqaures-estimates}. $\square$

\begin{lemma}
\label{lemma:grad-estimation}
Let $f: D \subset \R^d \to \R $ be of class $C^\mu$,  $a \in D$, and $\B(a) \subset D$
be some neighborhood of a. Suppose 
around point $a$ we have $m$ neighboring points $v_k$, $ k = 1, \ldots, m$ 
with a, $v_1, \ldots, v_m \in \B(a) \subset D$.
Suppose further that 
$\frac
    {\partial^\mu f}
    {\partial a_1^{l_1} \ldots \partial a_d^{l_d}}
    \in Lip_{\vartheta_{\max}}(\B(a))
$
for $l_1 + \ldots + l_d = \mu$, and we approximate the gradient locally at $a$
by $E_1 \hat \omega$ via the least-squares solution
$\hat \omega = \arg \min_{\omega \in \R^d} || A\omega - q||$,
where
\begin{equation}
    A = \begin{pmatrix}
        \nu_1^T & \nu_{1*}^T \\
        \nu_2^T & \nu_{2*}^T \\
        \vdots               \\
        \nu_m^T & \nu_{m*}^T \\
    \end{pmatrix} \in \R^{m\times d}, 
    \quad \quad
    q = \begin{pmatrix}
        \frac{f(a + h_1 \nu_1) - f(a)}{h_1} \\
        \frac{f(a + h_2 \nu_2) - f(a)}{h_1} \\
        \vdots \\
        \frac{f(a + h_m \nu_m) - f(a)}{h_m} \\
    \end{pmatrix} 
    \in \R^{mx1},
\end{equation}
$E_1 = \begin{pmatrix}
    I_d 0 
\end{pmatrix}
\in \R^{d\times p}
$, 
$ h_k = || v_k - a || $
with 
$ h_k \nu_k = v_k - a$;
$ \nu_k^T = (\nu_{1_k}, \ldots, \nu_{d_k}) $,

$ \nu_{k*}^T = 
\begin{pmatrix}
    \frac{h_k^{\mu'-1}}{\mu'!} 
    \binom{\mu'}{l_1, \ldots, l_d} 
    \prod_{i_1}^d \nu_i^{k_i}
\end{pmatrix}_{2 \le \mu' \le \mu, l_1 + \ldots l_d = \mu'}
$

and $
p = \sum_{i=1}^{\mu} \frac{(d+\mu)!}{d!}
$.
Then a bound on the error in the least-squares gradient estimate is given by:
\begin{equation}
    || \nabla f(a) - E_1 \hat \omega || \le \frac{\vartheta_{\max} h_{\max}^\mu}{\sigma_1 (\mu+1)! } \sqrt{\sum_{i=1}^m ||\nu_i||_1^{2\mu}},
\end{equation}
where $\sigma_1$ is the smallest singular value of A, which is assumed to have $\text{rank}(A) = p$,
$\vartheta_{\max}$ is as defined in Lemma \ref{lemma:least-sqaures-estimates},
and $h_{\max} = \max_{1\le k \le m} h_k$.
\end{lemma}

\emph{Proof.}
Let $E_2 \in \R^{(p-d)\times p}$ 
be the last $p-d$ rows of the identity matrix $I_p$
and 
\begin{equation}
U = \left(
\frac{\partial f}{\partial a_1}(a), \ldots,  \frac{\partial f}{\partial a_d}(a),
\frac{\partial^2 f}{
    \partial a_1^{l_{2_1}}
    \ldots
    \partial a_d^{l_{2_d}}
}
, \ldots,  
\frac{\partial^2 f}{\partial a_d^{l_{2_d}}}, 
\ldots,
\frac{\partial^\mu f}{\partial a_1^{l_{n_1}}}, \ldots, 
\frac{\partial^\mu f}{\partial a_d^{l_{n_d}}}
\right)^T \in \R^{p\times 1}.
\end{equation}

Now $U - \hat \omega$ can be partitioned as $
\begin{pmatrix}
E_1 ( U - \hat \omega ) \\
E_2 ( U - \hat \omega )
\end{pmatrix}
$,
and hence:
\begin{equation}
    || U - \hat \omega ||^2 =
    || E_1(U - \hat \omega) ||^2 + || E_2 (U - \hat \omega) ||^2 
    \ge 
    || E_1(U - \hat \omega) ||^2  
    = || \nabla f(a) - E_1 \hat \omega ||^2.
\end{equation}
Next, with $\hat \omega = A^\dagger q$ and $||A^\dagger|| = \frac{1}{\sigma_1^2}$, we have
\begin{equation}
    || U - \hat \omega ||^2 
    = || U - A^\dagger q ||^2 
    = || A^\dagger ( A U - q ) ||^2 
    \le || A^\dagger ||^2  || A U - q ||^2 
    = \frac{1}{\sigma_1^2} || A U - q ||^2.
\end{equation}
Now using the result in Lemma \ref{lemma:least-sqaures-estimates}, 
the following upper bound can be derived:
\begin{align}
    &|| A U - q ||^2  = 
        \sum_{i=1}^m \left|\left| 
            \sum_{k=1}^p A_{i,k} U_k - q_k
        \right|\right|^2 \\
    &= \sum_{i=1}^m \left|\left| \sum_{k=1}^p A_{i,k} U_k - q_k\right|\right|^2 \\
    &= \sum_{i=1}^m \left|\left| 
        \sum^\mu_{k=1} 
            \frac{h_i^{k-1}}{k!} 
                (\nu_i \cdot \nabla)^k f(a) 
                - \frac{f(a + h \nu) - f(a)}{h} 
    \right|\right|^2 \\
    & \le 
        \sum_{i=1}^m 
            \left(
                \frac{h_i^\mu}{(\mu+1)!} 
                    \vartheta_{\max}  ||\nu_i||_1^\mu
            \right)^2
            \quad \quad \text{(using Lemma \ref{lemma:least-sqaures-estimates})} \\
    & =
        \left(
            \frac{\vartheta_{\max}}{(\mu+1)!}
        \right)^2
        \sum_{i=1}^m 
            \left(
                h_i^\mu ||\nu_i||_1^\mu
            \right)^2
        \\
    & \le
        \left(
            \frac{\vartheta_{\max}}{(\mu+1)!}
        \right)^2
        \left(
            h_{\max}^\mu
        \right)^2
        \sum_{i=1}^m ||\nu_i||_1^{2\mu}
\end{align}

The result follows from
\begin{equation}
    ||\nabla f(a) - E_1 \hat \omega ||
    \le || U - \hat \omega ||
    \le 
        \frac{\vartheta_{\max} h_{\max}^\mu}
              {\sigma_1 (\mu+1)! } 
             \sqrt{\sum_{i=1}^m ||\nu_i||_1^{2\mu}}.
    \quad
    \square
\end{equation}

\subsection{DaNN pointwise error}
\label{appendix:pointwise_error}

The proof follows the one from \cite{Chen2018ExplainingTS}.
In \cite{Chen2018ExplainingTS}, also a method to break ties, e.g. points with the same distance is proposed -- a common problem when the data is discrete and not continuous. 
We will not discuss how to break ties, as the gradient estimation assumes continuous dimensions where ties should a.s. never happen.

\fakeparagraph{Technical Assumptions ($\Atechnical$):}
\begin{itemize}
    \item The feature space $\mathcal{X}$ and distance $\rho$
    form a separable metric space.
    \item The feature distribution $\mathbb{P}_X$ is a Borel probability measure.
\end{itemize}

\fakeparagraph{Assumptions Besicovitch($\Abesicovitch$):}
    The regression function $\eta$ satisfied the Besicovitch 
    condition if
    $\lim_{r \downarrow 0} \E[Y | X \in \B_{x, r}] = \eta(x)$
    for $x$ almost everywhere w.r.t. $\mathbb{P}_X$..

\fakeparagraph{Assumption Lipschitz ($\Alipschitz$):}
    The regression function $\eta$ is Lipschitz
    continuous with parameter $\vartheta_{\max}$ if
    $|\eta(x) - \eta(x') | \le \vartheta_{\max} \rho(x, x') $
    for all $x, x' \in \mathcal{X}$.

\begin{lemma}
\label{lemma:DNNR_pointwise_error}
Under assumptions $\Atechnical$, $\Abesicovitch$, 
let 
$x \in \text{supp}(\P_X)$ be a feature vector, 
and $\eta(x) = \E[Y | X = x] \in \R$,
be the expected label value for x.
Let  $\varepsilon > 0$ be an error tolerance 
in estimating expected label $\eta(x)$,
and $\delta \in (0, 1)$ be a probability tolerance.
Suppose that $Y \in [y_{\min}, y_{\max}]$ for some 
constants $y_{\min}$ and $y_{\max}$.
Let $\xi \in (0, 1)$. 
Then there exists a threshold distance $h^* \in (0, \inf)$
such that for any smaller distance $h \in (0, h^*)$
and with the number of nearest neighbors satisfying 
$k \le (1 - \xi) n \P_X(\Bxh)$, 
then with probability at least
\begin{equation}
    1 - 2 \exp\left(
        - \frac
            {k \varepsilon^2}
            {2 (y_{\max} - y_{\min})^2}
    \right)
    - 
    \exp\left(
        - \frac{\xi^2 n \P_X(\B_{x, r}}{2}
    \right).
\end{equation}
we have
\begin{equation}
    |\dnnlong(x) - \eta(x)| \le \varepsilon.
\end{equation}
Furthermore, if the function $\eta$ satisfies assumptions
$\Alipschitz$, then we can take 
\begin{equation}
\label{eq:h_star_dnn}
     h_{\text{DNNR}}^* = \sqrt{
        \frac
            {\varepsilon }
            {\vartheta_{\max}
            \left(
                1 
                + 
                \tau
            \right)
            }
    },
\end{equation}
where $
    \tau\!=\!\E \left[
            \frac{
                  \sqrt{\sum_{i=1}^m ||\nu_i||_1^{2\mu}}
            }
                  {\sigma_1 } 
            \:\big|\:  X \in  \Bxh 
        \right] 
$.
$\nu$, $\sigma_1$,  and 
$\vartheta_{\max}$ are defined as in Lemma \ref{lemma:grad_estimation}.

\end{lemma}

\emph{Proof of Lemma \ref{lemma:DNNR_pointwise_error}:}
Fix $x \in \text{supp}(\P_X)$. Let $\varepsilon > 0$.
We upper-bound the error $|\hat \eta - \eta(x)|$ with the 
triangle inequality:
\begin{align*}
    |\hat \eta(x) - \eta(x)| 
        &= |
            \hat \eta(x) 
            - \E_n[\dnn(x)]
            + \E_n[\dnn(x)]
            -  \eta(x)
        | \\
        &\le 
        \underbrace{|
            \hat \eta(x) 
            - \E_n[\dnn(x)]
        |}_{\text{\circled{1}}
        } + 
        \underbrace{|
            \E_n[\dnn(x)]
            -  \eta(x)
        |}_{\text{\circled{2}}}
        \\
\end{align*}
The proof now continues by showing that both \circled{1} and \circled{2}
are below $ \varepsilon / 2$ with high probability. The proof is adapted from the KNN pointwise regression proof in  \citep[p.~68ff.]{Chen2018ExplainingTS}.
The part \circled{1} is almost identical to the proof in \citep[p.~68ff.]{Chen2018ExplainingTS}. 
For part \circled{2}, we will use the Taylor Approximation and then bound the gradient using the results from Lemma \ref{lemma:grad_estimation}.

\fakeparagraph{Part \circled{1}  $|\dnn(x) - \E_n[\dnn(x)]| \le \frac{\varepsilon}{ 2}$:}

\begin{lemma}
\label{lemma:estimate_using_n_points} 
    Under same assumption of Theorem \ref{theorem:DNNR_pointwise_error},
    we have:
    \begin{equation}
        \P\left(
            |\dnn(x) - \E_n[\dnn(x)]|  \ge  \frac{\varepsilon}{2}
        \right) 
        \le 
        2 \exp\left(
            - \frac
                {k \varepsilon^2}
                {2 (y_{\max} - y_{\min})^2}
        \right).
    \end{equation}
\end{lemma}

\emph{Proof Lemma \ref{lemma:estimate_using_n_points}:}
As we want to apply the Hoeffding's inequality, we have to show 
that the $\hat Y$ are independent 

\fakeparagraph{Probability Model:}
The randomness of the $\dnnlong(x) $
can be described as:
\begin{enumerate}
    \item Sample a feature vector $\tilde X \in \mathcal{X}$ 
    from the marginal distribution of the $(k+1)$-st nearest neighbor
    of $x$, let $h_{k+1} = \rho(x, \tilde X)$ denote the distance between $x$ and $\tilde X$.
    \item Sample $k$ feature vectors i.i.d. from $\P_X$ conditioned on
    landing in the ball $\B^o_{x, \rho(x, \tilde X)}$,
    \item Sample $n - k - 1$ feature vectors i.i.d. from $\P_X$ conditoned on landing in $\mathcal{X} \setminus \B_{x, h_{k+1}}^o $,
    \item Randomly permute the $n$ feature vectors sampled,
    \item For each feature vector $X_i$ generated, sample its label 
    $Y_i$ based on the conditional distribution $\P_{Y|X=X_i}$.
\end{enumerate}
Therefore, we can write the expectation over the $n$ training points as:
\begin{equation}
    \label{eq:expectation_eta_n}
    \E_n[\dnn(x)] = \E_{h_{k+1}(x)}[
        \E[\hat Y|X \in \B_{x, h_{k+1}}^o
    ],
\end{equation}
where $\hat Y = Y + \nabla Y (X - x)$ denotes the prediction for point $x$ 
from a data point $X$ with label $Y$ and gradient $\nabla Y$.

The points samples in step 2 are precisely the k nearest neighbor of $x$,
and their $\hat Y$ values i.i.d. with expectation 
$\E_n[\dnn(x)] = \E_{\tilde X}[\E[\hat Y| X \in \B^o_{x, h_{k+1}}]$.
As they are bounded between 
$ y_{\min} $  and  $ y_{\max}$ (this can be enforced by clipping), Hoeffding's inequality yields:
\begin{equation}
    \P\left(
        |\dnn(x) - \E_n[\dnn(x)]|  \ge  \frac{\varepsilon}{2}
    \right) 
    \le 
    2 \exp\left(
        - \frac
            {k \varepsilon^2}
            {2 (y_{\max} - y_{\min})^2}
    \right)
\end{equation}
This finishes the proof of Lemma \ref{lemma:estimate_using_n_points}. $\quad \square$

\fakeparagraph{Part \circled{2}  $|\E_n[\dnn(x)] - \eta(x)|
\le \frac{\varepsilon}{ 2}$:}

As discussed above \eqref{eq:expectation_eta_n} the expectation of $\dnn(x)$ is
\begin{equation}
    \E_n[\dnn(x)] = \E_{h_{k+1}(x)}[E[Y|X \in \B_{x, h_{k+1}}^o]
\end{equation}
Suppose that we could show that there exists home 
$h > 0$ such that
\begin{equation}
    \label{eq:suppose-h-exists}
    |\E[Y| X \in \B^o_{x, r}] - \eta(x) | 
    \le \frac{\varepsilon}{2}
    \quad 
    \quad 
    \text{for all } r \in (0, h].
\end{equation}
Then provided that $h_{k+1} \le h$:
\begin{align}
    |\E_n[\dnn(x) - \eta(x)]| &= 
        |\E_{h_{k+1}(x)}[E[Y|X \in \B_{x, h_{k+1}}^o] - \eta(x) ]|
        \\
    &\le 
        \E_{h_{k+1}(x)}[|E[Y|X \in \B_{x, h_{k+1}}^o] - \eta(x) | ]
        \quad
        & \text{(Jensen's ineq.)} 
        \\
    &\le  \, \frac{\varepsilon}{2}        
        & \text{(inequality \eqref{eq:suppose-h-exists})}  
    \label{ineq:if_good_h} 
\end{align}
Before establishing the existence of $h$, we first show that for any
distance $r > 0$, with high probability we can ensure 
that $h_{k+1} \le r$. Thus, once we show that $h$ exists,
we also know that we can ensure that $h_{k+1} \le h$  with high probability.
\begin{lemma}
\label{lemma:k+1-within-r}.
Let $r > 0$ and $\xi \in (0, 1)$. 
For positive integer $k \le (1 - \xi) n \P_X(\B_{x, r})$,
\begin{equation}
    \P_X(h_{k+1}(x) \ge r)  \le 
        \exp\left(
            - \frac{\xi^2 n \P_X(\B_{x, r}}{2}
        \right).
\end{equation}
\end{lemma}
Thus, we have 
    $h_{k+1}(x) \le r $ with probability at least 
    $ 1 - \exp\left(
            - 0.5 \xi^2 n \P_X(\B_{x, r})
        \right)
    $.
\emph{Proof of Lemma \ref{lemma:k+1-within-r}.}
Fix $r > 0$ and $\xi \in (0, 1)$. 
Let $N_{x, r}$ be the number of training points
that land in the closed ball $\B_{x,r}$.
Note that 
$N_{x, r} \sim \text{Binomial}(n, \P_X(\B_{x, r}))$.
Then by a Chernoff bound for the binomial distribution, 
for any integer $k \le (1 - \xi) n \P_X(\B_{x, r})$, we have
\begin{align}
    \P(N_{x,r} \le k ) 
    &\le \exp\left(
        - \frac
            {(n \P_X(\B_{x, r} - k)^2}
            {2n \P_X(\B_{x, r})}
    \right)
    \\
    &\le \exp\left(
        - \frac
            {(n \P_X(\B_{x, r} - (1 - \xi) n \P_X(\B_{x, r}))^2}
            {2n \P_X(\B_{x, r})}
    \right) \\
    &= \exp\left(
        - \frac
            {\xi^2 n \P_X(\B_{x, r}))}
            {2}
    \right).
\end{align}
If $N_{x,r} \le k $, then also $ h_{k+1}(x) \ge r$.
Therefore, the event $
\{h_{k+1}(x) \ge r \}
\subset \{N_{x,r} \le k \}
$ and we have:
\begin{equation}
    \P_X(h_{k+1}(x) \ge r)  \le 
    \P(N_{x,r} \le k ) 
    \le \exp\left(
        - \frac
            {\xi^2 n \P_X(\B_{x, r}))}
            {2}
    \right).  
    \quad \square
\end{equation}

Now, we show which distance $h$ ensures that inequality \eqref{eq:suppose-h-exists} holds.
When we only know that
\begin{equation}
    \lim_{r\downarrow0} \E[Y|X \in \B^o_{x, r}] = \eta(x),
\end{equation}
then the definition of a limit implies that there exists $h^* > 0$
(that depends on $x$ and $\varepsilon$) such that
\begin{equation}
    |\E[Y | X \in \B^o_{x,h}] - \eta(x) | \le \frac{\varepsilon}{2} 
    \quad
    \text{for all }
    h \in (0, h^*),
\end{equation}
i.e., inequality \eqref{eq:suppose-h-exists} holds, and so we have 
$ 
    |\E_n[\dnn(x)] - \eta(x)| \le \frac{\varepsilon}{ 2}
$ as shown earlier in inequality \eqref{ineq:if_good_h}.

In the following derivation, we will 
assume $\eta$ to be Lipschitz continuous with parameters $\vartheta_{\max}$.
Further, we move $\eta(x)$ inside the expectation
and use it's first term of Taylor series 
with $X$ as root point: $\eta(\vx) = \eta(X) + \eta'(X) (\vx - X) + o(|\vx - X|)$, where $o(|\vx - X|)$ bounds the higher-order terms.

\def\Exyy{\E}    
\begin{align*}
    \big|
        \Exyy & \left[
             \dnn(\vx)
             \:|\:  X \in  \Bxh 
        \right] 
        - \eta(\vx) 
    \big|
    = \\
    &= \left| 
        \Exyy \left[
            Y + \nabla Y (\vx - X) 
            \:|\:  X \in  \Bxh 
        \right] 
        - \eta(\vx) 
    \right|  \\
    &= \left| 
        \Exyy \left[
            Y + \nabla Y (\vx - X) 
            - \eta(X) - \eta'(X) (\vx - X)
            + o(|x - X|)
            \:|\:  X \in  \Bxh 
        \right] 
    \right|  \\
    &= \left| 
        \Exyy \left[
            Y - \eta(X) 
            + \left(\nabla Y  - \eta'(X) \right) (\vx - X)
            + o(|x - X|)
            \:|\:  X \in  \Bxh 
        \right] 
    \right|  \\
\end{align*}
Now, we know that
$
    \Exyy[
            |Y - \eta(X)|
            \:|\:  X \in  \Bxh 
        ] = 0
$, as the noise term has zero mean.
\begin{align*}
    &= \left| 
        \Exyy\left[
            \left(\nabla Y  - \eta'(X) \right) (\vx - X)
            + o(|x - X|)
            \:|\:  X \in  \Bxh 
        \right] 
    \right|  \\
    & \le 
        \Exyy\left[
            \left| 
                \left(\nabla Y  - \eta'(X) \right) (\vx - X)
            \right|  
            +
            \left| 
                o(|x - X|)
            \right|  
            \:\big|\:  X \in  \Bxh 
        \right] 
    \\
\end{align*}
We can bound $ |x - X| < h_{\max}$ and $|\nabla Y - \eta'(X)|$ 
by using the results from Lemma \ref{lemma:least-sqaures-estimates}.
The higher order terms $o(|x - X|)$ can be bound by the remainder of the Talyor series:
$|R_\mu| \le 
    \frac
        { \vartheta_\mu |X - x|^\mu}
        { (\mu + 1)! }
$, where $\vartheta_\mu \le \vartheta_{\max}$. 

\begin{align}
    \Exyy & \left[
            \left| 
                \left(\nabla Y  - \eta'(X) \right) (\vx - X)
            \right|  
            +
            \left| 
                o(|x - X|)
            \right|  
            \:\big|\:  X \in  \Bxh 
        \right] 
    \\
    & \le
    \Exyy \left[
            \frac{\vartheta_{\max} h_{\max}^{\mu+1}}
                  {\sigma_1 (\mu+1)! } 
                  \sqrt{\sum_{i=1}^m ||\nu_i||_1^{2\mu}}
            + \vartheta_{\max} \frac{h_{\max}^2}{2}
            \:\Bigg|\:  X \in  \Bxh 
        \right] 
    \\
    & \le 
        \frac{\vartheta_{\max} h_{\max}^{\mu+1}}
              {(\mu+1)! } 
        \;
        \Exyy \left[
                \frac{
                      \sqrt{\sum_{i=1}^m ||\nu_i||_1^{2\mu}}
                }
                      {\sigma_1 } 
                \:\Bigg|\:  X \in  \Bxh 
            \right] 
        + 
        \vartheta_{\max} \frac{h_{\max}^2}{2} 
    \\
    & \le
        \frac{ \tau \vartheta_{\max} h_{\max}^{\mu+1} }
              {(\mu+1)! } 
        + 
        \vartheta_{\max} \frac{h_{\max}^2}{2} 
    \le \frac{\varepsilon}{2}
\end{align}
We used in the last step $
    \tau = \Exyy \left[
            \frac{
                  \sqrt{\sum_{i=1}^m ||\nu_i||_1^{2\mu}}
            }
                  {\sigma_1 } 
            \:\bigg|\:  X \in  \Bxh 
        \right] 
$. As this proof only concerns first-order approximations, we use $\mu = 1$:
\begin{align}
    & \frac{
        \tau \vartheta_{\max} h_{\max}^2}
              {2} 
        + 
        \vartheta_{\max} \frac{h_{\max}^2}{2} 
    \le 
    \frac{\varepsilon}{2} \\
    &
    \Leftrightarrow  
    h_{\max} \le \sqrt{
    \frac{\varepsilon}{\vartheta_{\max} (1 + \tau)}
    }
\end{align}
This finishes the proof of Lemma \ref{lemma:DNNR_pointwise_error} 
$\square$.

Theorem \ref{theorem:DNNR_pointwise_error} follows from selecting
$\xi = \frac{1}{2}$ and observing that:

\begin{align}
    n \ge \frac{8}{\P_X(\Bxh)} \log \frac{2}{\delta} 
    &\Rightarrow 
    \exp \left(
        -\frac{\xi^2 n \P_X(\Bxh)}
                {2}
    \right)
    \le \frac{\delta}{2}, \\
    k \ge \frac{2(y_{\max} - y_{\min})^2}{\varepsilon^2} \log \frac{4}{\delta} 
    & \Rightarrow 
    \exp \left(
        -\frac{k \varepsilon^2}
            { 2(y_{\max} - y_{\min})^2 }
    \right)
    \le \frac{\delta}{2}.
\end{align}

\section{Hyperparameters}
\label{appendix:hyperparams}
\begin{table}[H]
\centering
\caption{Number of Hyperparameters tuned for each model on the benchmark datasets. For DNNR, KNN and MLP, we take small datasets to be $n<2000$, and medium datasets to be $n<50000$. The same applies for PMLB and Feynman. We were unable to tune TabNet model due to computation constraints and 
used their Tabnet-L configuration for larger datasets (Sarcos, Protein, CO$^2$ and NOx Emissions) and Tabnet-S (n\_d and n\_a = 16) for smaller datasets.}
\label{tab:number_hyperparams}
\begin{tabular}{lrrrrrrr}
\toprule
{} & Airfoil & CO$^2$Emission & California & Concrete & NOxEmission & Protein & Yacht \\
\midrule
DNNR                   &            150 &                 40 &                       40 &             150 &                 40 &             40 &          150 \\
LL                     &             50 &                 50 &                       50 &              50 &                 50 &             50 &           50 \\
Random Forest          &             12 &                 12 &                       12 &              12 &                 12 &             12 &           12 \\
Grad. B. Trees &             48 &                 48 &                       48 &              48 &                 48 &             48 &           48 \\
MLP                    &            594 &                 72 &                       72 &             594 &                 72 &             72 &          594 \\
CatBoost               &             48 &                 48 &                       48 &              48 &                 48 &             48 &           48 \\
XGBoost                &             48 &                 48 &                       48 &              48 &                 48 &             48 &           48 \\
LGBM                   &             48 &                 48 &                       48 &              48 &                 48 &             48 &           48 \\
KNN                    &            384 &                 64 &                       64 &             384 &                 64 &             64 &          384 \\
Tabnet                 &            1 &                 1 &                       1 &             1 &                 1 &             1 &          1 \\

\bottomrule
\end{tabular}

\end{table}
\begin{table}[H]
\centering
\caption{XGBoost Hyperparameters}
\begin{tabular}{lll}
\toprule
       learning\_rate & max\_depth &      n\_estimators \\
\midrule
{[0.001,0.01,0.1,0.3]} &  {[3,5,10]} & {[50,100,500,1000]} \\
\bottomrule
\end{tabular}

\end{table}
\begin{table}[H]
\centering
\caption{LightGBM Hyperparameters}
\begin{tabular}{lll}
\toprule
       learning\_rate & max\_depth &      n\_estimators \\
\midrule
{[0.001,0.01,0.1,0.3]} &  {[3,5,10]} & {[50,100,500,1000]} \\
\bottomrule
\end{tabular}

\end{table}
\begin{table}[H]
\centering
\caption{CatBoost Hyperparameters}
\begin{tabular}{llll}
\toprule
verbose &        learning\_rate & max\_depth &      n\_estimators \\
\midrule
{[False]} & {[0.001,0.01,0.1,0.3]} &  {[3,5,10]} & {[50,100,500,1000]} \\
\bottomrule
\end{tabular}

\end{table}
\begin{table}[H]
\centering
\caption{Gradient Boosting Hyperparameters}
\begin{tabular}{lll}
\toprule
       learning\_rate & max\_depth &      n\_estimators \\
\midrule
{[0.001,0.01,0.1,0.3]} &  {[3,5,10]} & {[50,100,500,1000]} \\
\bottomrule
\end{tabular}

\end{table}
\begin{table}[H]
\centering
\caption{Random Forests Hyper Parameters}
\begin{tabular}{lll}
\toprule
criterion &      n\_estimators &     max\_features \\
\midrule
    {[mse]} & {[50,100,500,1000]} & {[auto,sqrt,log2]} \\
\bottomrule
\end{tabular}

\end{table}
\begin{table}[H]
\centering
\caption{MLP Hyperparameters for small datasets}
\begin{tabular}{l}
\toprule
                                hidden\_layer\_sizes \\
\midrule
{[(25,), (50,), (100,), (250,),
        (25, 25), (50, 50), (100, 100), (250, 250),
        (25, 25, 25), (50, 50, 50), (100, 100, 100)}  \\
\bottomrule
\end{tabular}

\begin{tabular}{lllll}
\toprule
      alpha &  batch\_size &                  learning\_rate & learning\_rate\_init & early\_stopping \\
\midrule
  {[0,0.01,1]} & {[64,128]} & {[constant,invscaling,adaptive]} &   {[0.001,0.01,0.1]} &         {[True]} \\
\bottomrule
\end{tabular}

\end{table}
\begin{table}[H]
\centering
\caption{MLP Hyperparameters for medium datasets}
\begin{tabular}{ll}
\toprule
                 hidden\_layer\_sizes &    alpha \\
\midrule
{[(128,),(128, 128),(128, 128, 128)]} & {[0,0.01]} \\
\bottomrule
\end{tabular}

\begin{tabular}{llll}
\toprule
batch\_size &                  learning\_rate &      learning\_rate\_init & early\_stopping \\
\midrule
     {[512]} & {[constant,invscaling,adaptive]} & {[0.0001,0.001,0.01,0.1]} &         {[True]} \\
\bottomrule
\end{tabular}

\end{table}
\begin{table}[H]
\centering
\caption{MLP Hyperparameters for large datasets}
\begin{tabular}{ll}
\toprule
                 hidden\_layer\_sizes & alpha \\
\midrule
{[(128,),(128, 128),(128, 128, 128)]} &   {[0]} \\
\bottomrule
\end{tabular}

\begin{tabular}{llll}
\toprule
batch\_size &       learning\_rate &  learning\_rate\_init & early\_stopping \\
\midrule
     {[512]} & {[constant,adaptive]} & {[0.0001,0.001,0.01]} &         {[True]} \\
\bottomrule
\end{tabular}

\end{table}
\begin{table}[H]
\centering
\caption{KNN Hyperparameters for small datasets}
\begin{tabular}{ll}
\toprule
           n\_neighbors &            weights \\
\midrule
{[2,5,7,10,20,30,40,50]} & {[uniform,distance]} \\
\bottomrule
\end{tabular}

\begin{tabular}{lll}
\toprule
                algorithm &      leaf\_size &     p \\
\midrule
{[ball\_tree,kd\_tree,brute]} & {[10,30,50,100]} & {[1,2]} \\
\bottomrule
\end{tabular}

\end{table}
\begin{table}[H]
\centering
\caption{KNN Hyperparameters for medium datasets}
\begin{tabular}{llll}
\toprule
             n\_neighbors &            weights &      leaf\_size &   p \\
\midrule
{[2,5,7,10,25,50,100,250]} & {[distance,uniform]} & {[10,30,50,100]} & {[2]} \\
\bottomrule
\end{tabular}

\end{table}
\begin{table}[H]
\centering
\caption{KNN Hyperparameters for large datasets}
\begin{tabular}{llll}
\toprule
             n\_neighbors &    weights &      leaf\_size &   p \\
\midrule
{[2,3,5,7,10,12,15,20,25]} & {[distance]} & {[10,30,50,100]} & {[2]} \\
\bottomrule
\end{tabular}

\end{table}
\begin{table}[H]
\centering
\caption{Tabnet Hyperparameters for large datasets}
\begin{tabular}{llllll}
\toprule
  n\_d & verbose &   n\_a & lambda\_sparse & batch\_size & virtual\_batch\_size \\
\midrule
{[128]} & {[False]} & {[128]} &      {[0.0001]} &     {[4096]} &              {[128]} \\
\bottomrule
\end{tabular}

\begin{tabular}{lllllll}
\toprule
momentum & n\_steps & gamma & optimizer\_params &                    scheduler\_params & max\_epochs & patience \\
\midrule
   {[0.8]} &     {[5]} & {[1.5]} &   {[\{'lr': 0.02\}]} & {[\{'step\_size': 8000, 'gamma': 0.9\}]} &     {[3000]} &    {[100]} \\
\bottomrule
\end{tabular}

\end{table}
\begin{table}[H]
\centering
\caption{Tabnet Hyperparameters for small datasets}
\begin{tabular}{llllll}
\toprule
   n\_d &    n\_a & verbose & lambda\_sparse & batch\_size & virtual\_batch\_size \\
\midrule
{[8,16]} & {[8,16]} & {[False]} &      {[0.0001]} &   {[64,128]} &             {[8,16]} \\
\bottomrule
\end{tabular}

\begin{tabular}{lllllll}
\toprule
momentum & n\_steps & gamma & optimizer\_params &                   scheduler\_params & max\_epochs & patience \\
\midrule
  {[0.02]} &     {[3]} & {[1.3]} &   {[\{'lr': 0.02\}]} & {[\{'step\_size': 10, 'gamma': 0.95\}]} &     {[3000]} &    {[100]} \\
\bottomrule
\end{tabular}

\end{table}
\begin{table}[H]
\centering
\caption{DNNR Hyperparameters for small datasets, $n\_neighbohrs$ corresponds to the $k$. For the $k'$ (number of neighbors used in approximating the gradient) we use n values sampled between $lower\_bound \times d$ and $lower\_bound \times d$}
\begin{tabular}{llll}
\toprule
n\_neighbhors & upper\_bound & lower\_bound &    n \\
\midrule
   {[1, 2, 3, 5, 7]} &        {[15]} &         {[2]} & {[30]} \\
\bottomrule
\end{tabular}

\end{table}
\begin{table}[H]
\centering
\caption{DNNR Hyperparameters for medium datasets , $n\_neighbohrs$ corresponds to the $k$. For the $k'$ (number of neighbors used in approximating the gradient) we use n values sampled between $lower\_bound \times d$ and $lower\_bound \times d$}
\begin{tabular}{llll}
\toprule
n\_neighbhors & upper\_bound & lower\_bound &    n \\
\midrule
       {[3,4]} &        {[18]} &         {[2]} & {[20]} \\
\bottomrule
\end{tabular}

\end{table}
\begin{table}[H]
\centering
\caption{DNNR Hyperparameters for large datasets , $n\_neighbohrs$ corresponds to the $k$. For the $k'$ (number of neighbors used in approximating the gradient) we use n values sampled between $lower\_bound \times d$ and $lower\_bound \times d$}
\begin{tabular}{llll}
\toprule
n\_neighbhors & upper\_bound & lower\_bound &    n \\
\midrule
         {[3]} &        {[12]} &         {[2]} & {[14]} \\
\bottomrule
\end{tabular}

\end{table}
\begin{table}[H]
\centering
\caption{LL Hyperparameters  $n\_neighbohrs$ corresponds to the $k$. For the size of the neighborhood we use n values sampled between $lower\_bound \times d$ and $lower\_bound \times d$}
\begin{tabular}{lll}
\toprule
upper\_bound & lower\_bound &    n \\
\midrule
       {[25]} &         {[2]} & {[50]} \\
\bottomrule
\end{tabular}

\end{table}

\end{document}

%% file: math_commands.tex
\usepackage{amsmath,amsfonts,bm}

\def\1{\bm{1}}

\def\vx{{\bm{x}}}

\DeclareMathAlphabet{\mathsfit}{\encodingdefault}{\sfdefault}{m}{sl}
\SetMathAlphabet{\mathsfit}{bold}{\encodingdefault}{\sfdefault}{bx}{n}

\newcommand{\E}{\mathbb{E}}

\newcommand{\R}{\mathbb{R}}



%% file: leons_defs.tex
\newcommand{\missingcite}[1]{{\color{red}(Missing et al., 2030)}}
\newcommand{\fakeparagraph}[1]{\textbf{#1}\hspace{0.3em}}

\newcommand*\circled[1]{\tikz[baseline=(char.base)]{
    \node[shape=circle,draw,inner sep=1pt] (char) {#1};}}

%% file: figures/bounds/error_tolerance.pgf
\begingroup%
\makeatletter%
\begin{pgfpicture}%
\pgfpathrectangle{\pgfpointorigin}{\pgfqpoint{2.560444in}{1.266874in}}%
\pgfusepath{use as bounding box, clip}%
\begin{pgfscope}%
\pgfsetbuttcap%
\pgfsetmiterjoin%
\definecolor{currentfill}{rgb}{1.000000,1.000000,1.000000}%
\pgfsetfillcolor{currentfill}%
\pgfsetlinewidth{0.000000pt}%
\definecolor{currentstroke}{rgb}{1.000000,1.000000,1.000000}%
\pgfsetstrokecolor{currentstroke}%
\pgfsetstrokeopacity{0.000000}%
\pgfsetdash{}{0pt}%
\pgfpathmoveto{\pgfqpoint{0.000000in}{-0.000000in}}%
\pgfpathlineto{\pgfqpoint{2.560444in}{-0.000000in}}%
\pgfpathlineto{\pgfqpoint{2.560444in}{1.266874in}}%
\pgfpathlineto{\pgfqpoint{0.000000in}{1.266874in}}%
\pgfpathclose%
\pgfusepath{fill}%
\end{pgfscope}%
\begin{pgfscope}%
\pgfsetbuttcap%
\pgfsetmiterjoin%
\definecolor{currentfill}{rgb}{1.000000,1.000000,1.000000}%
\pgfsetfillcolor{currentfill}%
\pgfsetlinewidth{0.000000pt}%
\definecolor{currentstroke}{rgb}{0.000000,0.000000,0.000000}%
\pgfsetstrokecolor{currentstroke}%
\pgfsetstrokeopacity{0.000000}%
\pgfsetdash{}{0pt}%
\pgfpathmoveto{\pgfqpoint{0.467944in}{0.349549in}}%
\pgfpathlineto{\pgfqpoint{2.560444in}{0.349549in}}%
\pgfpathlineto{\pgfqpoint{2.560444in}{1.266874in}}%
\pgfpathlineto{\pgfqpoint{0.467944in}{1.266874in}}%
\pgfpathclose%
\pgfusepath{fill}%
\end{pgfscope}%
\begin{pgfscope}%
\pgfsetbuttcap%
\pgfsetroundjoin%
\definecolor{currentfill}{rgb}{0.150000,0.150000,0.150000}%
\pgfsetfillcolor{currentfill}%
\pgfsetlinewidth{0.501875pt}%
\definecolor{currentstroke}{rgb}{0.150000,0.150000,0.150000}%
\pgfsetstrokecolor{currentstroke}%
\pgfsetdash{}{0pt}%
\pgfsys@defobject{currentmarker}{\pgfqpoint{0.000000in}{-0.048611in}}{\pgfqpoint{0.000000in}{0.000000in}}{%
\pgfpathmoveto{\pgfqpoint{0.000000in}{0.000000in}}%
\pgfpathlineto{\pgfqpoint{0.000000in}{-0.048611in}}%
\pgfusepath{stroke,fill}%
}%
\begin{pgfscope}%
\pgfsys@transformshift{0.563058in}{0.349549in}%
\pgfsys@useobject{currentmarker}{}%
\end{pgfscope}%
\end{pgfscope}%
\begin{pgfscope}%
\definecolor{textcolor}{rgb}{0.150000,0.150000,0.150000}%
\pgfsetstrokecolor{textcolor}%
\pgfsetfillcolor{textcolor}%
\pgftext[x=0.563058in,y=0.252327in,,top]{\color{textcolor}\sffamily\fontsize{7.000000}{8.400000}\selectfont \(\displaystyle {10^{-3}}\)}%
\end{pgfscope}%
\begin{pgfscope}%
\pgfsetbuttcap%
\pgfsetroundjoin%
\definecolor{currentfill}{rgb}{0.150000,0.150000,0.150000}%
\pgfsetfillcolor{currentfill}%
\pgfsetlinewidth{0.501875pt}%
\definecolor{currentstroke}{rgb}{0.150000,0.150000,0.150000}%
\pgfsetstrokecolor{currentstroke}%
\pgfsetdash{}{0pt}%
\pgfsys@defobject{currentmarker}{\pgfqpoint{0.000000in}{-0.048611in}}{\pgfqpoint{0.000000in}{0.000000in}}{%
\pgfpathmoveto{\pgfqpoint{0.000000in}{0.000000in}}%
\pgfpathlineto{\pgfqpoint{0.000000in}{-0.048611in}}%
\pgfusepath{stroke,fill}%
}%
\begin{pgfscope}%
\pgfsys@transformshift{1.197149in}{0.349549in}%
\pgfsys@useobject{currentmarker}{}%
\end{pgfscope}%
\end{pgfscope}%
\begin{pgfscope}%
\definecolor{textcolor}{rgb}{0.150000,0.150000,0.150000}%
\pgfsetstrokecolor{textcolor}%
\pgfsetfillcolor{textcolor}%
\pgftext[x=1.197149in,y=0.252327in,,top]{\color{textcolor}\sffamily\fontsize{7.000000}{8.400000}\selectfont \(\displaystyle {10^{-2}}\)}%
\end{pgfscope}%
\begin{pgfscope}%
\pgfsetbuttcap%
\pgfsetroundjoin%
\definecolor{currentfill}{rgb}{0.150000,0.150000,0.150000}%
\pgfsetfillcolor{currentfill}%
\pgfsetlinewidth{0.501875pt}%
\definecolor{currentstroke}{rgb}{0.150000,0.150000,0.150000}%
\pgfsetstrokecolor{currentstroke}%
\pgfsetdash{}{0pt}%
\pgfsys@defobject{currentmarker}{\pgfqpoint{0.000000in}{-0.048611in}}{\pgfqpoint{0.000000in}{0.000000in}}{%
\pgfpathmoveto{\pgfqpoint{0.000000in}{0.000000in}}%
\pgfpathlineto{\pgfqpoint{0.000000in}{-0.048611in}}%
\pgfusepath{stroke,fill}%
}%
\begin{pgfscope}%
\pgfsys@transformshift{1.831239in}{0.349549in}%
\pgfsys@useobject{currentmarker}{}%
\end{pgfscope}%
\end{pgfscope}%
\begin{pgfscope}%
\definecolor{textcolor}{rgb}{0.150000,0.150000,0.150000}%
\pgfsetstrokecolor{textcolor}%
\pgfsetfillcolor{textcolor}%
\pgftext[x=1.831239in,y=0.252327in,,top]{\color{textcolor}\sffamily\fontsize{7.000000}{8.400000}\selectfont \(\displaystyle {10^{-1}}\)}%
\end{pgfscope}%
\begin{pgfscope}%
\pgfsetbuttcap%
\pgfsetroundjoin%
\definecolor{currentfill}{rgb}{0.150000,0.150000,0.150000}%
\pgfsetfillcolor{currentfill}%
\pgfsetlinewidth{0.501875pt}%
\definecolor{currentstroke}{rgb}{0.150000,0.150000,0.150000}%
\pgfsetstrokecolor{currentstroke}%
\pgfsetdash{}{0pt}%
\pgfsys@defobject{currentmarker}{\pgfqpoint{0.000000in}{-0.048611in}}{\pgfqpoint{0.000000in}{0.000000in}}{%
\pgfpathmoveto{\pgfqpoint{0.000000in}{0.000000in}}%
\pgfpathlineto{\pgfqpoint{0.000000in}{-0.048611in}}%
\pgfusepath{stroke,fill}%
}%
\begin{pgfscope}%
\pgfsys@transformshift{2.465330in}{0.349549in}%
\pgfsys@useobject{currentmarker}{}%
\end{pgfscope}%
\end{pgfscope}%
\begin{pgfscope}%
\definecolor{textcolor}{rgb}{0.150000,0.150000,0.150000}%
\pgfsetstrokecolor{textcolor}%
\pgfsetfillcolor{textcolor}%
\pgftext[x=2.465330in,y=0.252327in,,top]{\color{textcolor}\sffamily\fontsize{7.000000}{8.400000}\selectfont \(\displaystyle {10^{0}}\)}%
\end{pgfscope}%
\begin{pgfscope}%
\pgfsetbuttcap%
\pgfsetroundjoin%
\definecolor{currentfill}{rgb}{0.150000,0.150000,0.150000}%
\pgfsetfillcolor{currentfill}%
\pgfsetlinewidth{0.501875pt}%
\definecolor{currentstroke}{rgb}{0.150000,0.150000,0.150000}%
\pgfsetstrokecolor{currentstroke}%
\pgfsetdash{}{0pt}%
\pgfsys@defobject{currentmarker}{\pgfqpoint{0.000000in}{-0.027778in}}{\pgfqpoint{0.000000in}{0.000000in}}{%
\pgfpathmoveto{\pgfqpoint{0.000000in}{0.000000in}}%
\pgfpathlineto{\pgfqpoint{0.000000in}{-0.027778in}}%
\pgfusepath{stroke,fill}%
}%
\begin{pgfscope}%
\pgfsys@transformshift{0.501608in}{0.349549in}%
\pgfsys@useobject{currentmarker}{}%
\end{pgfscope}%
\end{pgfscope}%
\begin{pgfscope}%
\pgfsetbuttcap%
\pgfsetroundjoin%
\definecolor{currentfill}{rgb}{0.150000,0.150000,0.150000}%
\pgfsetfillcolor{currentfill}%
\pgfsetlinewidth{0.501875pt}%
\definecolor{currentstroke}{rgb}{0.150000,0.150000,0.150000}%
\pgfsetstrokecolor{currentstroke}%
\pgfsetdash{}{0pt}%
\pgfsys@defobject{currentmarker}{\pgfqpoint{0.000000in}{-0.027778in}}{\pgfqpoint{0.000000in}{0.000000in}}{%
\pgfpathmoveto{\pgfqpoint{0.000000in}{0.000000in}}%
\pgfpathlineto{\pgfqpoint{0.000000in}{-0.027778in}}%
\pgfusepath{stroke,fill}%
}%
\begin{pgfscope}%
\pgfsys@transformshift{0.534043in}{0.349549in}%
\pgfsys@useobject{currentmarker}{}%
\end{pgfscope}%
\end{pgfscope}%
\begin{pgfscope}%
\pgfsetbuttcap%
\pgfsetroundjoin%
\definecolor{currentfill}{rgb}{0.150000,0.150000,0.150000}%
\pgfsetfillcolor{currentfill}%
\pgfsetlinewidth{0.501875pt}%
\definecolor{currentstroke}{rgb}{0.150000,0.150000,0.150000}%
\pgfsetstrokecolor{currentstroke}%
\pgfsetdash{}{0pt}%
\pgfsys@defobject{currentmarker}{\pgfqpoint{0.000000in}{-0.027778in}}{\pgfqpoint{0.000000in}{0.000000in}}{%
\pgfpathmoveto{\pgfqpoint{0.000000in}{0.000000in}}%
\pgfpathlineto{\pgfqpoint{0.000000in}{-0.027778in}}%
\pgfusepath{stroke,fill}%
}%
\begin{pgfscope}%
\pgfsys@transformshift{0.753938in}{0.349549in}%
\pgfsys@useobject{currentmarker}{}%
\end{pgfscope}%
\end{pgfscope}%
\begin{pgfscope}%
\pgfsetbuttcap%
\pgfsetroundjoin%
\definecolor{currentfill}{rgb}{0.150000,0.150000,0.150000}%
\pgfsetfillcolor{currentfill}%
\pgfsetlinewidth{0.501875pt}%
\definecolor{currentstroke}{rgb}{0.150000,0.150000,0.150000}%
\pgfsetstrokecolor{currentstroke}%
\pgfsetdash{}{0pt}%
\pgfsys@defobject{currentmarker}{\pgfqpoint{0.000000in}{-0.027778in}}{\pgfqpoint{0.000000in}{0.000000in}}{%
\pgfpathmoveto{\pgfqpoint{0.000000in}{0.000000in}}%
\pgfpathlineto{\pgfqpoint{0.000000in}{-0.027778in}}%
\pgfusepath{stroke,fill}%
}%
\begin{pgfscope}%
\pgfsys@transformshift{0.865596in}{0.349549in}%
\pgfsys@useobject{currentmarker}{}%
\end{pgfscope}%
\end{pgfscope}%
\begin{pgfscope}%
\pgfsetbuttcap%
\pgfsetroundjoin%
\definecolor{currentfill}{rgb}{0.150000,0.150000,0.150000}%
\pgfsetfillcolor{currentfill}%
\pgfsetlinewidth{0.501875pt}%
\definecolor{currentstroke}{rgb}{0.150000,0.150000,0.150000}%
\pgfsetstrokecolor{currentstroke}%
\pgfsetdash{}{0pt}%
\pgfsys@defobject{currentmarker}{\pgfqpoint{0.000000in}{-0.027778in}}{\pgfqpoint{0.000000in}{0.000000in}}{%
\pgfpathmoveto{\pgfqpoint{0.000000in}{0.000000in}}%
\pgfpathlineto{\pgfqpoint{0.000000in}{-0.027778in}}%
\pgfusepath{stroke,fill}%
}%
\begin{pgfscope}%
\pgfsys@transformshift{0.944818in}{0.349549in}%
\pgfsys@useobject{currentmarker}{}%
\end{pgfscope}%
\end{pgfscope}%
\begin{pgfscope}%
\pgfsetbuttcap%
\pgfsetroundjoin%
\definecolor{currentfill}{rgb}{0.150000,0.150000,0.150000}%
\pgfsetfillcolor{currentfill}%
\pgfsetlinewidth{0.501875pt}%
\definecolor{currentstroke}{rgb}{0.150000,0.150000,0.150000}%
\pgfsetstrokecolor{currentstroke}%
\pgfsetdash{}{0pt}%
\pgfsys@defobject{currentmarker}{\pgfqpoint{0.000000in}{-0.027778in}}{\pgfqpoint{0.000000in}{0.000000in}}{%
\pgfpathmoveto{\pgfqpoint{0.000000in}{0.000000in}}%
\pgfpathlineto{\pgfqpoint{0.000000in}{-0.027778in}}%
\pgfusepath{stroke,fill}%
}%
\begin{pgfscope}%
\pgfsys@transformshift{1.006268in}{0.349549in}%
\pgfsys@useobject{currentmarker}{}%
\end{pgfscope}%
\end{pgfscope}%
\begin{pgfscope}%
\pgfsetbuttcap%
\pgfsetroundjoin%
\definecolor{currentfill}{rgb}{0.150000,0.150000,0.150000}%
\pgfsetfillcolor{currentfill}%
\pgfsetlinewidth{0.501875pt}%
\definecolor{currentstroke}{rgb}{0.150000,0.150000,0.150000}%
\pgfsetstrokecolor{currentstroke}%
\pgfsetdash{}{0pt}%
\pgfsys@defobject{currentmarker}{\pgfqpoint{0.000000in}{-0.027778in}}{\pgfqpoint{0.000000in}{0.000000in}}{%
\pgfpathmoveto{\pgfqpoint{0.000000in}{0.000000in}}%
\pgfpathlineto{\pgfqpoint{0.000000in}{-0.027778in}}%
\pgfusepath{stroke,fill}%
}%
\begin{pgfscope}%
\pgfsys@transformshift{1.056476in}{0.349549in}%
\pgfsys@useobject{currentmarker}{}%
\end{pgfscope}%
\end{pgfscope}%
\begin{pgfscope}%
\pgfsetbuttcap%
\pgfsetroundjoin%
\definecolor{currentfill}{rgb}{0.150000,0.150000,0.150000}%
\pgfsetfillcolor{currentfill}%
\pgfsetlinewidth{0.501875pt}%
\definecolor{currentstroke}{rgb}{0.150000,0.150000,0.150000}%
\pgfsetstrokecolor{currentstroke}%
\pgfsetdash{}{0pt}%
\pgfsys@defobject{currentmarker}{\pgfqpoint{0.000000in}{-0.027778in}}{\pgfqpoint{0.000000in}{0.000000in}}{%
\pgfpathmoveto{\pgfqpoint{0.000000in}{0.000000in}}%
\pgfpathlineto{\pgfqpoint{0.000000in}{-0.027778in}}%
\pgfusepath{stroke,fill}%
}%
\begin{pgfscope}%
\pgfsys@transformshift{1.098927in}{0.349549in}%
\pgfsys@useobject{currentmarker}{}%
\end{pgfscope}%
\end{pgfscope}%
\begin{pgfscope}%
\pgfsetbuttcap%
\pgfsetroundjoin%
\definecolor{currentfill}{rgb}{0.150000,0.150000,0.150000}%
\pgfsetfillcolor{currentfill}%
\pgfsetlinewidth{0.501875pt}%
\definecolor{currentstroke}{rgb}{0.150000,0.150000,0.150000}%
\pgfsetstrokecolor{currentstroke}%
\pgfsetdash{}{0pt}%
\pgfsys@defobject{currentmarker}{\pgfqpoint{0.000000in}{-0.027778in}}{\pgfqpoint{0.000000in}{0.000000in}}{%
\pgfpathmoveto{\pgfqpoint{0.000000in}{0.000000in}}%
\pgfpathlineto{\pgfqpoint{0.000000in}{-0.027778in}}%
\pgfusepath{stroke,fill}%
}%
\begin{pgfscope}%
\pgfsys@transformshift{1.135699in}{0.349549in}%
\pgfsys@useobject{currentmarker}{}%
\end{pgfscope}%
\end{pgfscope}%
\begin{pgfscope}%
\pgfsetbuttcap%
\pgfsetroundjoin%
\definecolor{currentfill}{rgb}{0.150000,0.150000,0.150000}%
\pgfsetfillcolor{currentfill}%
\pgfsetlinewidth{0.501875pt}%
\definecolor{currentstroke}{rgb}{0.150000,0.150000,0.150000}%
\pgfsetstrokecolor{currentstroke}%
\pgfsetdash{}{0pt}%
\pgfsys@defobject{currentmarker}{\pgfqpoint{0.000000in}{-0.027778in}}{\pgfqpoint{0.000000in}{0.000000in}}{%
\pgfpathmoveto{\pgfqpoint{0.000000in}{0.000000in}}%
\pgfpathlineto{\pgfqpoint{0.000000in}{-0.027778in}}%
\pgfusepath{stroke,fill}%
}%
\begin{pgfscope}%
\pgfsys@transformshift{1.168134in}{0.349549in}%
\pgfsys@useobject{currentmarker}{}%
\end{pgfscope}%
\end{pgfscope}%
\begin{pgfscope}%
\pgfsetbuttcap%
\pgfsetroundjoin%
\definecolor{currentfill}{rgb}{0.150000,0.150000,0.150000}%
\pgfsetfillcolor{currentfill}%
\pgfsetlinewidth{0.501875pt}%
\definecolor{currentstroke}{rgb}{0.150000,0.150000,0.150000}%
\pgfsetstrokecolor{currentstroke}%
\pgfsetdash{}{0pt}%
\pgfsys@defobject{currentmarker}{\pgfqpoint{0.000000in}{-0.027778in}}{\pgfqpoint{0.000000in}{0.000000in}}{%
\pgfpathmoveto{\pgfqpoint{0.000000in}{0.000000in}}%
\pgfpathlineto{\pgfqpoint{0.000000in}{-0.027778in}}%
\pgfusepath{stroke,fill}%
}%
\begin{pgfscope}%
\pgfsys@transformshift{1.388029in}{0.349549in}%
\pgfsys@useobject{currentmarker}{}%
\end{pgfscope}%
\end{pgfscope}%
\begin{pgfscope}%
\pgfsetbuttcap%
\pgfsetroundjoin%
\definecolor{currentfill}{rgb}{0.150000,0.150000,0.150000}%
\pgfsetfillcolor{currentfill}%
\pgfsetlinewidth{0.501875pt}%
\definecolor{currentstroke}{rgb}{0.150000,0.150000,0.150000}%
\pgfsetstrokecolor{currentstroke}%
\pgfsetdash{}{0pt}%
\pgfsys@defobject{currentmarker}{\pgfqpoint{0.000000in}{-0.027778in}}{\pgfqpoint{0.000000in}{0.000000in}}{%
\pgfpathmoveto{\pgfqpoint{0.000000in}{0.000000in}}%
\pgfpathlineto{\pgfqpoint{0.000000in}{-0.027778in}}%
\pgfusepath{stroke,fill}%
}%
\begin{pgfscope}%
\pgfsys@transformshift{1.499687in}{0.349549in}%
\pgfsys@useobject{currentmarker}{}%
\end{pgfscope}%
\end{pgfscope}%
\begin{pgfscope}%
\pgfsetbuttcap%
\pgfsetroundjoin%
\definecolor{currentfill}{rgb}{0.150000,0.150000,0.150000}%
\pgfsetfillcolor{currentfill}%
\pgfsetlinewidth{0.501875pt}%
\definecolor{currentstroke}{rgb}{0.150000,0.150000,0.150000}%
\pgfsetstrokecolor{currentstroke}%
\pgfsetdash{}{0pt}%
\pgfsys@defobject{currentmarker}{\pgfqpoint{0.000000in}{-0.027778in}}{\pgfqpoint{0.000000in}{0.000000in}}{%
\pgfpathmoveto{\pgfqpoint{0.000000in}{0.000000in}}%
\pgfpathlineto{\pgfqpoint{0.000000in}{-0.027778in}}%
\pgfusepath{stroke,fill}%
}%
\begin{pgfscope}%
\pgfsys@transformshift{1.578909in}{0.349549in}%
\pgfsys@useobject{currentmarker}{}%
\end{pgfscope}%
\end{pgfscope}%
\begin{pgfscope}%
\pgfsetbuttcap%
\pgfsetroundjoin%
\definecolor{currentfill}{rgb}{0.150000,0.150000,0.150000}%
\pgfsetfillcolor{currentfill}%
\pgfsetlinewidth{0.501875pt}%
\definecolor{currentstroke}{rgb}{0.150000,0.150000,0.150000}%
\pgfsetstrokecolor{currentstroke}%
\pgfsetdash{}{0pt}%
\pgfsys@defobject{currentmarker}{\pgfqpoint{0.000000in}{-0.027778in}}{\pgfqpoint{0.000000in}{0.000000in}}{%
\pgfpathmoveto{\pgfqpoint{0.000000in}{0.000000in}}%
\pgfpathlineto{\pgfqpoint{0.000000in}{-0.027778in}}%
\pgfusepath{stroke,fill}%
}%
\begin{pgfscope}%
\pgfsys@transformshift{1.640359in}{0.349549in}%
\pgfsys@useobject{currentmarker}{}%
\end{pgfscope}%
\end{pgfscope}%
\begin{pgfscope}%
\pgfsetbuttcap%
\pgfsetroundjoin%
\definecolor{currentfill}{rgb}{0.150000,0.150000,0.150000}%
\pgfsetfillcolor{currentfill}%
\pgfsetlinewidth{0.501875pt}%
\definecolor{currentstroke}{rgb}{0.150000,0.150000,0.150000}%
\pgfsetstrokecolor{currentstroke}%
\pgfsetdash{}{0pt}%
\pgfsys@defobject{currentmarker}{\pgfqpoint{0.000000in}{-0.027778in}}{\pgfqpoint{0.000000in}{0.000000in}}{%
\pgfpathmoveto{\pgfqpoint{0.000000in}{0.000000in}}%
\pgfpathlineto{\pgfqpoint{0.000000in}{-0.027778in}}%
\pgfusepath{stroke,fill}%
}%
\begin{pgfscope}%
\pgfsys@transformshift{1.690567in}{0.349549in}%
\pgfsys@useobject{currentmarker}{}%
\end{pgfscope}%
\end{pgfscope}%
\begin{pgfscope}%
\pgfsetbuttcap%
\pgfsetroundjoin%
\definecolor{currentfill}{rgb}{0.150000,0.150000,0.150000}%
\pgfsetfillcolor{currentfill}%
\pgfsetlinewidth{0.501875pt}%
\definecolor{currentstroke}{rgb}{0.150000,0.150000,0.150000}%
\pgfsetstrokecolor{currentstroke}%
\pgfsetdash{}{0pt}%
\pgfsys@defobject{currentmarker}{\pgfqpoint{0.000000in}{-0.027778in}}{\pgfqpoint{0.000000in}{0.000000in}}{%
\pgfpathmoveto{\pgfqpoint{0.000000in}{0.000000in}}%
\pgfpathlineto{\pgfqpoint{0.000000in}{-0.027778in}}%
\pgfusepath{stroke,fill}%
}%
\begin{pgfscope}%
\pgfsys@transformshift{1.733018in}{0.349549in}%
\pgfsys@useobject{currentmarker}{}%
\end{pgfscope}%
\end{pgfscope}%
\begin{pgfscope}%
\pgfsetbuttcap%
\pgfsetroundjoin%
\definecolor{currentfill}{rgb}{0.150000,0.150000,0.150000}%
\pgfsetfillcolor{currentfill}%
\pgfsetlinewidth{0.501875pt}%
\definecolor{currentstroke}{rgb}{0.150000,0.150000,0.150000}%
\pgfsetstrokecolor{currentstroke}%
\pgfsetdash{}{0pt}%
\pgfsys@defobject{currentmarker}{\pgfqpoint{0.000000in}{-0.027778in}}{\pgfqpoint{0.000000in}{0.000000in}}{%
\pgfpathmoveto{\pgfqpoint{0.000000in}{0.000000in}}%
\pgfpathlineto{\pgfqpoint{0.000000in}{-0.027778in}}%
\pgfusepath{stroke,fill}%
}%
\begin{pgfscope}%
\pgfsys@transformshift{1.769790in}{0.349549in}%
\pgfsys@useobject{currentmarker}{}%
\end{pgfscope}%
\end{pgfscope}%
\begin{pgfscope}%
\pgfsetbuttcap%
\pgfsetroundjoin%
\definecolor{currentfill}{rgb}{0.150000,0.150000,0.150000}%
\pgfsetfillcolor{currentfill}%
\pgfsetlinewidth{0.501875pt}%
\definecolor{currentstroke}{rgb}{0.150000,0.150000,0.150000}%
\pgfsetstrokecolor{currentstroke}%
\pgfsetdash{}{0pt}%
\pgfsys@defobject{currentmarker}{\pgfqpoint{0.000000in}{-0.027778in}}{\pgfqpoint{0.000000in}{0.000000in}}{%
\pgfpathmoveto{\pgfqpoint{0.000000in}{0.000000in}}%
\pgfpathlineto{\pgfqpoint{0.000000in}{-0.027778in}}%
\pgfusepath{stroke,fill}%
}%
\begin{pgfscope}%
\pgfsys@transformshift{1.802225in}{0.349549in}%
\pgfsys@useobject{currentmarker}{}%
\end{pgfscope}%
\end{pgfscope}%
\begin{pgfscope}%
\pgfsetbuttcap%
\pgfsetroundjoin%
\definecolor{currentfill}{rgb}{0.150000,0.150000,0.150000}%
\pgfsetfillcolor{currentfill}%
\pgfsetlinewidth{0.501875pt}%
\definecolor{currentstroke}{rgb}{0.150000,0.150000,0.150000}%
\pgfsetstrokecolor{currentstroke}%
\pgfsetdash{}{0pt}%
\pgfsys@defobject{currentmarker}{\pgfqpoint{0.000000in}{-0.027778in}}{\pgfqpoint{0.000000in}{0.000000in}}{%
\pgfpathmoveto{\pgfqpoint{0.000000in}{0.000000in}}%
\pgfpathlineto{\pgfqpoint{0.000000in}{-0.027778in}}%
\pgfusepath{stroke,fill}%
}%
\begin{pgfscope}%
\pgfsys@transformshift{2.022120in}{0.349549in}%
\pgfsys@useobject{currentmarker}{}%
\end{pgfscope}%
\end{pgfscope}%
\begin{pgfscope}%
\pgfsetbuttcap%
\pgfsetroundjoin%
\definecolor{currentfill}{rgb}{0.150000,0.150000,0.150000}%
\pgfsetfillcolor{currentfill}%
\pgfsetlinewidth{0.501875pt}%
\definecolor{currentstroke}{rgb}{0.150000,0.150000,0.150000}%
\pgfsetstrokecolor{currentstroke}%
\pgfsetdash{}{0pt}%
\pgfsys@defobject{currentmarker}{\pgfqpoint{0.000000in}{-0.027778in}}{\pgfqpoint{0.000000in}{0.000000in}}{%
\pgfpathmoveto{\pgfqpoint{0.000000in}{0.000000in}}%
\pgfpathlineto{\pgfqpoint{0.000000in}{-0.027778in}}%
\pgfusepath{stroke,fill}%
}%
\begin{pgfscope}%
\pgfsys@transformshift{2.133778in}{0.349549in}%
\pgfsys@useobject{currentmarker}{}%
\end{pgfscope}%
\end{pgfscope}%
\begin{pgfscope}%
\pgfsetbuttcap%
\pgfsetroundjoin%
\definecolor{currentfill}{rgb}{0.150000,0.150000,0.150000}%
\pgfsetfillcolor{currentfill}%
\pgfsetlinewidth{0.501875pt}%
\definecolor{currentstroke}{rgb}{0.150000,0.150000,0.150000}%
\pgfsetstrokecolor{currentstroke}%
\pgfsetdash{}{0pt}%
\pgfsys@defobject{currentmarker}{\pgfqpoint{0.000000in}{-0.027778in}}{\pgfqpoint{0.000000in}{0.000000in}}{%
\pgfpathmoveto{\pgfqpoint{0.000000in}{0.000000in}}%
\pgfpathlineto{\pgfqpoint{0.000000in}{-0.027778in}}%
\pgfusepath{stroke,fill}%
}%
\begin{pgfscope}%
\pgfsys@transformshift{2.213000in}{0.349549in}%
\pgfsys@useobject{currentmarker}{}%
\end{pgfscope}%
\end{pgfscope}%
\begin{pgfscope}%
\pgfsetbuttcap%
\pgfsetroundjoin%
\definecolor{currentfill}{rgb}{0.150000,0.150000,0.150000}%
\pgfsetfillcolor{currentfill}%
\pgfsetlinewidth{0.501875pt}%
\definecolor{currentstroke}{rgb}{0.150000,0.150000,0.150000}%
\pgfsetstrokecolor{currentstroke}%
\pgfsetdash{}{0pt}%
\pgfsys@defobject{currentmarker}{\pgfqpoint{0.000000in}{-0.027778in}}{\pgfqpoint{0.000000in}{0.000000in}}{%
\pgfpathmoveto{\pgfqpoint{0.000000in}{0.000000in}}%
\pgfpathlineto{\pgfqpoint{0.000000in}{-0.027778in}}%
\pgfusepath{stroke,fill}%
}%
\begin{pgfscope}%
\pgfsys@transformshift{2.274450in}{0.349549in}%
\pgfsys@useobject{currentmarker}{}%
\end{pgfscope}%
\end{pgfscope}%
\begin{pgfscope}%
\pgfsetbuttcap%
\pgfsetroundjoin%
\definecolor{currentfill}{rgb}{0.150000,0.150000,0.150000}%
\pgfsetfillcolor{currentfill}%
\pgfsetlinewidth{0.501875pt}%
\definecolor{currentstroke}{rgb}{0.150000,0.150000,0.150000}%
\pgfsetstrokecolor{currentstroke}%
\pgfsetdash{}{0pt}%
\pgfsys@defobject{currentmarker}{\pgfqpoint{0.000000in}{-0.027778in}}{\pgfqpoint{0.000000in}{0.000000in}}{%
\pgfpathmoveto{\pgfqpoint{0.000000in}{0.000000in}}%
\pgfpathlineto{\pgfqpoint{0.000000in}{-0.027778in}}%
\pgfusepath{stroke,fill}%
}%
\begin{pgfscope}%
\pgfsys@transformshift{2.324658in}{0.349549in}%
\pgfsys@useobject{currentmarker}{}%
\end{pgfscope}%
\end{pgfscope}%
\begin{pgfscope}%
\pgfsetbuttcap%
\pgfsetroundjoin%
\definecolor{currentfill}{rgb}{0.150000,0.150000,0.150000}%
\pgfsetfillcolor{currentfill}%
\pgfsetlinewidth{0.501875pt}%
\definecolor{currentstroke}{rgb}{0.150000,0.150000,0.150000}%
\pgfsetstrokecolor{currentstroke}%
\pgfsetdash{}{0pt}%
\pgfsys@defobject{currentmarker}{\pgfqpoint{0.000000in}{-0.027778in}}{\pgfqpoint{0.000000in}{0.000000in}}{%
\pgfpathmoveto{\pgfqpoint{0.000000in}{0.000000in}}%
\pgfpathlineto{\pgfqpoint{0.000000in}{-0.027778in}}%
\pgfusepath{stroke,fill}%
}%
\begin{pgfscope}%
\pgfsys@transformshift{2.367108in}{0.349549in}%
\pgfsys@useobject{currentmarker}{}%
\end{pgfscope}%
\end{pgfscope}%
\begin{pgfscope}%
\pgfsetbuttcap%
\pgfsetroundjoin%
\definecolor{currentfill}{rgb}{0.150000,0.150000,0.150000}%
\pgfsetfillcolor{currentfill}%
\pgfsetlinewidth{0.501875pt}%
\definecolor{currentstroke}{rgb}{0.150000,0.150000,0.150000}%
\pgfsetstrokecolor{currentstroke}%
\pgfsetdash{}{0pt}%
\pgfsys@defobject{currentmarker}{\pgfqpoint{0.000000in}{-0.027778in}}{\pgfqpoint{0.000000in}{0.000000in}}{%
\pgfpathmoveto{\pgfqpoint{0.000000in}{0.000000in}}%
\pgfpathlineto{\pgfqpoint{0.000000in}{-0.027778in}}%
\pgfusepath{stroke,fill}%
}%
\begin{pgfscope}%
\pgfsys@transformshift{2.403881in}{0.349549in}%
\pgfsys@useobject{currentmarker}{}%
\end{pgfscope}%
\end{pgfscope}%
\begin{pgfscope}%
\pgfsetbuttcap%
\pgfsetroundjoin%
\definecolor{currentfill}{rgb}{0.150000,0.150000,0.150000}%
\pgfsetfillcolor{currentfill}%
\pgfsetlinewidth{0.501875pt}%
\definecolor{currentstroke}{rgb}{0.150000,0.150000,0.150000}%
\pgfsetstrokecolor{currentstroke}%
\pgfsetdash{}{0pt}%
\pgfsys@defobject{currentmarker}{\pgfqpoint{0.000000in}{-0.027778in}}{\pgfqpoint{0.000000in}{0.000000in}}{%
\pgfpathmoveto{\pgfqpoint{0.000000in}{0.000000in}}%
\pgfpathlineto{\pgfqpoint{0.000000in}{-0.027778in}}%
\pgfusepath{stroke,fill}%
}%
\begin{pgfscope}%
\pgfsys@transformshift{2.436316in}{0.349549in}%
\pgfsys@useobject{currentmarker}{}%
\end{pgfscope}%
\end{pgfscope}%
\begin{pgfscope}%
\definecolor{textcolor}{rgb}{0.150000,0.150000,0.150000}%
\pgfsetstrokecolor{textcolor}%
\pgfsetfillcolor{textcolor}%
\pgftext[x=1.514194in,y=0.101617in,,top]{\color{textcolor}\sffamily\fontsize{8.000000}{9.600000}\selectfont Error Tolerance [\(\displaystyle \varepsilon\)]}%
\end{pgfscope}%
\begin{pgfscope}%
\pgfsetbuttcap%
\pgfsetroundjoin%
\definecolor{currentfill}{rgb}{0.150000,0.150000,0.150000}%
\pgfsetfillcolor{currentfill}%
\pgfsetlinewidth{0.501875pt}%
\definecolor{currentstroke}{rgb}{0.150000,0.150000,0.150000}%
\pgfsetstrokecolor{currentstroke}%
\pgfsetdash{}{0pt}%
\pgfsys@defobject{currentmarker}{\pgfqpoint{-0.048611in}{0.000000in}}{\pgfqpoint{-0.000000in}{0.000000in}}{%
\pgfpathmoveto{\pgfqpoint{-0.000000in}{0.000000in}}%
\pgfpathlineto{\pgfqpoint{-0.048611in}{0.000000in}}%
\pgfusepath{stroke,fill}%
}%
\begin{pgfscope}%
\pgfsys@transformshift{0.467944in}{0.581698in}%
\pgfsys@useobject{currentmarker}{}%
\end{pgfscope}%
\end{pgfscope}%
\begin{pgfscope}%
\definecolor{textcolor}{rgb}{0.150000,0.150000,0.150000}%
\pgfsetstrokecolor{textcolor}%
\pgfsetfillcolor{textcolor}%
\pgftext[x=0.158528in, y=0.544553in, left, base]{\color{textcolor}\sffamily\fontsize{7.000000}{8.400000}\selectfont \(\displaystyle {10^{15}}\)}%
\end{pgfscope}%
\begin{pgfscope}%
\pgfsetbuttcap%
\pgfsetroundjoin%
\definecolor{currentfill}{rgb}{0.150000,0.150000,0.150000}%
\pgfsetfillcolor{currentfill}%
\pgfsetlinewidth{0.501875pt}%
\definecolor{currentstroke}{rgb}{0.150000,0.150000,0.150000}%
\pgfsetstrokecolor{currentstroke}%
\pgfsetdash{}{0pt}%
\pgfsys@defobject{currentmarker}{\pgfqpoint{-0.048611in}{0.000000in}}{\pgfqpoint{-0.000000in}{0.000000in}}{%
\pgfpathmoveto{\pgfqpoint{-0.000000in}{0.000000in}}%
\pgfpathlineto{\pgfqpoint{-0.048611in}{0.000000in}}%
\pgfusepath{stroke,fill}%
}%
\begin{pgfscope}%
\pgfsys@transformshift{0.467944in}{0.820318in}%
\pgfsys@useobject{currentmarker}{}%
\end{pgfscope}%
\end{pgfscope}%
\begin{pgfscope}%
\definecolor{textcolor}{rgb}{0.150000,0.150000,0.150000}%
\pgfsetstrokecolor{textcolor}%
\pgfsetfillcolor{textcolor}%
\pgftext[x=0.158528in, y=0.783173in, left, base]{\color{textcolor}\sffamily\fontsize{7.000000}{8.400000}\selectfont \(\displaystyle {10^{22}}\)}%
\end{pgfscope}%
\begin{pgfscope}%
\pgfsetbuttcap%
\pgfsetroundjoin%
\definecolor{currentfill}{rgb}{0.150000,0.150000,0.150000}%
\pgfsetfillcolor{currentfill}%
\pgfsetlinewidth{0.501875pt}%
\definecolor{currentstroke}{rgb}{0.150000,0.150000,0.150000}%
\pgfsetstrokecolor{currentstroke}%
\pgfsetdash{}{0pt}%
\pgfsys@defobject{currentmarker}{\pgfqpoint{-0.048611in}{0.000000in}}{\pgfqpoint{-0.000000in}{0.000000in}}{%
\pgfpathmoveto{\pgfqpoint{-0.000000in}{0.000000in}}%
\pgfpathlineto{\pgfqpoint{-0.048611in}{0.000000in}}%
\pgfusepath{stroke,fill}%
}%
\begin{pgfscope}%
\pgfsys@transformshift{0.467944in}{1.058939in}%
\pgfsys@useobject{currentmarker}{}%
\end{pgfscope}%
\end{pgfscope}%
\begin{pgfscope}%
\definecolor{textcolor}{rgb}{0.150000,0.150000,0.150000}%
\pgfsetstrokecolor{textcolor}%
\pgfsetfillcolor{textcolor}%
\pgftext[x=0.158528in, y=1.021794in, left, base]{\color{textcolor}\sffamily\fontsize{7.000000}{8.400000}\selectfont \(\displaystyle {10^{29}}\)}%
\end{pgfscope}%
\begin{pgfscope}%
\definecolor{textcolor}{rgb}{0.150000,0.150000,0.150000}%
\pgfsetstrokecolor{textcolor}%
\pgfsetfillcolor{textcolor}%
\pgftext[x=0.102973in,y=0.808212in,,bottom,rotate=90.000000]{\color{textcolor}\sffamily\fontsize{8.000000}{9.600000}\selectfont \#Samples [\(\displaystyle n\)]}%
\end{pgfscope}%
\begin{pgfscope}%
\pgfpathrectangle{\pgfqpoint{0.467944in}{0.349549in}}{\pgfqpoint{2.092500in}{0.917325in}}%
\pgfusepath{clip}%
\pgfsetroundcap%
\pgfsetroundjoin%
\pgfsetlinewidth{0.501875pt}%
\definecolor{currentstroke}{rgb}{0.003922,0.450980,0.698039}%
\pgfsetstrokecolor{currentstroke}%
\pgfsetdash{}{0pt}%
\pgfpathmoveto{\pgfqpoint{0.563058in}{0.851440in}}%
\pgfpathlineto{\pgfqpoint{0.582273in}{0.846792in}}%
\pgfpathlineto{\pgfqpoint{0.601487in}{0.842143in}}%
\pgfpathlineto{\pgfqpoint{0.620702in}{0.837495in}}%
\pgfpathlineto{\pgfqpoint{0.639917in}{0.832847in}}%
\pgfpathlineto{\pgfqpoint{0.659132in}{0.828198in}}%
\pgfpathlineto{\pgfqpoint{0.678347in}{0.823550in}}%
\pgfpathlineto{\pgfqpoint{0.697562in}{0.818901in}}%
\pgfpathlineto{\pgfqpoint{0.716777in}{0.814253in}}%
\pgfpathlineto{\pgfqpoint{0.735992in}{0.809604in}}%
\pgfpathlineto{\pgfqpoint{0.755206in}{0.804956in}}%
\pgfpathlineto{\pgfqpoint{0.774421in}{0.800307in}}%
\pgfpathlineto{\pgfqpoint{0.793636in}{0.795659in}}%
\pgfpathlineto{\pgfqpoint{0.812851in}{0.791010in}}%
\pgfpathlineto{\pgfqpoint{0.832066in}{0.786362in}}%
\pgfpathlineto{\pgfqpoint{0.851281in}{0.781714in}}%
\pgfpathlineto{\pgfqpoint{0.870496in}{0.777065in}}%
\pgfpathlineto{\pgfqpoint{0.889711in}{0.772417in}}%
\pgfpathlineto{\pgfqpoint{0.908925in}{0.767768in}}%
\pgfpathlineto{\pgfqpoint{0.928140in}{0.763120in}}%
\pgfpathlineto{\pgfqpoint{0.947355in}{0.758471in}}%
\pgfpathlineto{\pgfqpoint{0.966570in}{0.753823in}}%
\pgfpathlineto{\pgfqpoint{0.985785in}{0.749174in}}%
\pgfpathlineto{\pgfqpoint{1.005000in}{0.744526in}}%
\pgfpathlineto{\pgfqpoint{1.024215in}{0.739877in}}%
\pgfpathlineto{\pgfqpoint{1.043430in}{0.735229in}}%
\pgfpathlineto{\pgfqpoint{1.062644in}{0.730581in}}%
\pgfpathlineto{\pgfqpoint{1.081859in}{0.725932in}}%
\pgfpathlineto{\pgfqpoint{1.101074in}{0.721284in}}%
\pgfpathlineto{\pgfqpoint{1.120289in}{0.716635in}}%
\pgfpathlineto{\pgfqpoint{1.139504in}{0.711987in}}%
\pgfpathlineto{\pgfqpoint{1.158719in}{0.707338in}}%
\pgfpathlineto{\pgfqpoint{1.177934in}{0.702690in}}%
\pgfpathlineto{\pgfqpoint{1.197149in}{0.698041in}}%
\pgfpathlineto{\pgfqpoint{1.216363in}{0.693393in}}%
\pgfpathlineto{\pgfqpoint{1.235578in}{0.688744in}}%
\pgfpathlineto{\pgfqpoint{1.254793in}{0.684096in}}%
\pgfpathlineto{\pgfqpoint{1.274008in}{0.679448in}}%
\pgfpathlineto{\pgfqpoint{1.293223in}{0.674799in}}%
\pgfpathlineto{\pgfqpoint{1.312438in}{0.670151in}}%
\pgfpathlineto{\pgfqpoint{1.331653in}{0.665502in}}%
\pgfpathlineto{\pgfqpoint{1.350868in}{0.660854in}}%
\pgfpathlineto{\pgfqpoint{1.370082in}{0.656205in}}%
\pgfpathlineto{\pgfqpoint{1.389297in}{0.651557in}}%
\pgfpathlineto{\pgfqpoint{1.408512in}{0.646908in}}%
\pgfpathlineto{\pgfqpoint{1.427727in}{0.642260in}}%
\pgfpathlineto{\pgfqpoint{1.446942in}{0.637611in}}%
\pgfpathlineto{\pgfqpoint{1.466157in}{0.632963in}}%
\pgfpathlineto{\pgfqpoint{1.485372in}{0.628315in}}%
\pgfpathlineto{\pgfqpoint{1.504587in}{0.623666in}}%
\pgfpathlineto{\pgfqpoint{1.523801in}{0.619018in}}%
\pgfpathlineto{\pgfqpoint{1.543016in}{0.614369in}}%
\pgfpathlineto{\pgfqpoint{1.562231in}{0.609721in}}%
\pgfpathlineto{\pgfqpoint{1.581446in}{0.605072in}}%
\pgfpathlineto{\pgfqpoint{1.600661in}{0.600424in}}%
\pgfpathlineto{\pgfqpoint{1.619876in}{0.595775in}}%
\pgfpathlineto{\pgfqpoint{1.639091in}{0.591127in}}%
\pgfpathlineto{\pgfqpoint{1.658306in}{0.586478in}}%
\pgfpathlineto{\pgfqpoint{1.677520in}{0.581830in}}%
\pgfpathlineto{\pgfqpoint{1.696735in}{0.577182in}}%
\pgfpathlineto{\pgfqpoint{1.715950in}{0.572533in}}%
\pgfpathlineto{\pgfqpoint{1.735165in}{0.567885in}}%
\pgfpathlineto{\pgfqpoint{1.754380in}{0.563236in}}%
\pgfpathlineto{\pgfqpoint{1.773595in}{0.558588in}}%
\pgfpathlineto{\pgfqpoint{1.792810in}{0.553939in}}%
\pgfpathlineto{\pgfqpoint{1.812025in}{0.549291in}}%
\pgfpathlineto{\pgfqpoint{1.831239in}{0.544642in}}%
\pgfpathlineto{\pgfqpoint{1.850454in}{0.539994in}}%
\pgfpathlineto{\pgfqpoint{1.869669in}{0.535345in}}%
\pgfpathlineto{\pgfqpoint{1.888884in}{0.530697in}}%
\pgfpathlineto{\pgfqpoint{1.908099in}{0.526049in}}%
\pgfpathlineto{\pgfqpoint{1.927314in}{0.521400in}}%
\pgfpathlineto{\pgfqpoint{1.946529in}{0.516752in}}%
\pgfpathlineto{\pgfqpoint{1.965744in}{0.512103in}}%
\pgfpathlineto{\pgfqpoint{1.984958in}{0.507455in}}%
\pgfpathlineto{\pgfqpoint{2.004173in}{0.502806in}}%
\pgfpathlineto{\pgfqpoint{2.023388in}{0.498158in}}%
\pgfpathlineto{\pgfqpoint{2.042603in}{0.493509in}}%
\pgfpathlineto{\pgfqpoint{2.061818in}{0.488861in}}%
\pgfpathlineto{\pgfqpoint{2.081033in}{0.484212in}}%
\pgfpathlineto{\pgfqpoint{2.100248in}{0.479564in}}%
\pgfpathlineto{\pgfqpoint{2.119463in}{0.474916in}}%
\pgfpathlineto{\pgfqpoint{2.138677in}{0.470267in}}%
\pgfpathlineto{\pgfqpoint{2.157892in}{0.465619in}}%
\pgfpathlineto{\pgfqpoint{2.177107in}{0.460970in}}%
\pgfpathlineto{\pgfqpoint{2.196322in}{0.456322in}}%
\pgfpathlineto{\pgfqpoint{2.215537in}{0.451673in}}%
\pgfpathlineto{\pgfqpoint{2.234752in}{0.447025in}}%
\pgfpathlineto{\pgfqpoint{2.253967in}{0.442377in}}%
\pgfpathlineto{\pgfqpoint{2.273182in}{0.437728in}}%
\pgfpathlineto{\pgfqpoint{2.292397in}{0.433080in}}%
\pgfpathlineto{\pgfqpoint{2.311611in}{0.428431in}}%
\pgfpathlineto{\pgfqpoint{2.330826in}{0.423783in}}%
\pgfpathlineto{\pgfqpoint{2.350041in}{0.419135in}}%
\pgfpathlineto{\pgfqpoint{2.369256in}{0.414486in}}%
\pgfpathlineto{\pgfqpoint{2.388471in}{0.409838in}}%
\pgfpathlineto{\pgfqpoint{2.407686in}{0.405190in}}%
\pgfpathlineto{\pgfqpoint{2.426901in}{0.400541in}}%
\pgfpathlineto{\pgfqpoint{2.446116in}{0.395893in}}%
\pgfpathlineto{\pgfqpoint{2.465330in}{0.391246in}}%
\pgfusepath{stroke}%
\end{pgfscope}%
\begin{pgfscope}%
\pgfpathrectangle{\pgfqpoint{0.467944in}{0.349549in}}{\pgfqpoint{2.092500in}{0.917325in}}%
\pgfusepath{clip}%
\pgfsetroundcap%
\pgfsetroundjoin%
\pgfsetlinewidth{0.501875pt}%
\definecolor{currentstroke}{rgb}{0.870588,0.560784,0.019608}%
\pgfsetstrokecolor{currentstroke}%
\pgfsetdash{}{0pt}%
\pgfpathmoveto{\pgfqpoint{0.563058in}{1.225178in}}%
\pgfpathlineto{\pgfqpoint{0.582273in}{1.217947in}}%
\pgfpathlineto{\pgfqpoint{0.601487in}{1.210716in}}%
\pgfpathlineto{\pgfqpoint{0.620702in}{1.203485in}}%
\pgfpathlineto{\pgfqpoint{0.639917in}{1.196254in}}%
\pgfpathlineto{\pgfqpoint{0.659132in}{1.189023in}}%
\pgfpathlineto{\pgfqpoint{0.678347in}{1.181792in}}%
\pgfpathlineto{\pgfqpoint{0.697562in}{1.174561in}}%
\pgfpathlineto{\pgfqpoint{0.716777in}{1.167330in}}%
\pgfpathlineto{\pgfqpoint{0.735992in}{1.160099in}}%
\pgfpathlineto{\pgfqpoint{0.755206in}{1.152868in}}%
\pgfpathlineto{\pgfqpoint{0.774421in}{1.145637in}}%
\pgfpathlineto{\pgfqpoint{0.793636in}{1.138406in}}%
\pgfpathlineto{\pgfqpoint{0.812851in}{1.131175in}}%
\pgfpathlineto{\pgfqpoint{0.832066in}{1.123945in}}%
\pgfpathlineto{\pgfqpoint{0.851281in}{1.116714in}}%
\pgfpathlineto{\pgfqpoint{0.870496in}{1.109483in}}%
\pgfpathlineto{\pgfqpoint{0.889711in}{1.102252in}}%
\pgfpathlineto{\pgfqpoint{0.908925in}{1.095021in}}%
\pgfpathlineto{\pgfqpoint{0.928140in}{1.087790in}}%
\pgfpathlineto{\pgfqpoint{0.947355in}{1.080559in}}%
\pgfpathlineto{\pgfqpoint{0.966570in}{1.073328in}}%
\pgfpathlineto{\pgfqpoint{0.985785in}{1.066097in}}%
\pgfpathlineto{\pgfqpoint{1.005000in}{1.058866in}}%
\pgfpathlineto{\pgfqpoint{1.024215in}{1.051635in}}%
\pgfpathlineto{\pgfqpoint{1.043430in}{1.044404in}}%
\pgfpathlineto{\pgfqpoint{1.062644in}{1.037173in}}%
\pgfpathlineto{\pgfqpoint{1.081859in}{1.029942in}}%
\pgfpathlineto{\pgfqpoint{1.101074in}{1.022712in}}%
\pgfpathlineto{\pgfqpoint{1.120289in}{1.015481in}}%
\pgfpathlineto{\pgfqpoint{1.139504in}{1.008250in}}%
\pgfpathlineto{\pgfqpoint{1.158719in}{1.001019in}}%
\pgfpathlineto{\pgfqpoint{1.177934in}{0.993788in}}%
\pgfpathlineto{\pgfqpoint{1.197149in}{0.986557in}}%
\pgfpathlineto{\pgfqpoint{1.216363in}{0.979326in}}%
\pgfpathlineto{\pgfqpoint{1.235578in}{0.972095in}}%
\pgfpathlineto{\pgfqpoint{1.254793in}{0.964864in}}%
\pgfpathlineto{\pgfqpoint{1.274008in}{0.957633in}}%
\pgfpathlineto{\pgfqpoint{1.293223in}{0.950402in}}%
\pgfpathlineto{\pgfqpoint{1.312438in}{0.943171in}}%
\pgfpathlineto{\pgfqpoint{1.331653in}{0.935940in}}%
\pgfpathlineto{\pgfqpoint{1.350868in}{0.928709in}}%
\pgfpathlineto{\pgfqpoint{1.370082in}{0.921479in}}%
\pgfpathlineto{\pgfqpoint{1.389297in}{0.914248in}}%
\pgfpathlineto{\pgfqpoint{1.408512in}{0.907017in}}%
\pgfpathlineto{\pgfqpoint{1.427727in}{0.899786in}}%
\pgfpathlineto{\pgfqpoint{1.446942in}{0.892555in}}%
\pgfpathlineto{\pgfqpoint{1.466157in}{0.885324in}}%
\pgfpathlineto{\pgfqpoint{1.485372in}{0.878093in}}%
\pgfpathlineto{\pgfqpoint{1.504587in}{0.870862in}}%
\pgfpathlineto{\pgfqpoint{1.523801in}{0.863631in}}%
\pgfpathlineto{\pgfqpoint{1.543016in}{0.856400in}}%
\pgfpathlineto{\pgfqpoint{1.562231in}{0.849169in}}%
\pgfpathlineto{\pgfqpoint{1.581446in}{0.841938in}}%
\pgfpathlineto{\pgfqpoint{1.600661in}{0.834707in}}%
\pgfpathlineto{\pgfqpoint{1.619876in}{0.827476in}}%
\pgfpathlineto{\pgfqpoint{1.639091in}{0.820246in}}%
\pgfpathlineto{\pgfqpoint{1.658306in}{0.813015in}}%
\pgfpathlineto{\pgfqpoint{1.677520in}{0.805784in}}%
\pgfpathlineto{\pgfqpoint{1.696735in}{0.798553in}}%
\pgfpathlineto{\pgfqpoint{1.715950in}{0.791322in}}%
\pgfpathlineto{\pgfqpoint{1.735165in}{0.784091in}}%
\pgfpathlineto{\pgfqpoint{1.754380in}{0.776860in}}%
\pgfpathlineto{\pgfqpoint{1.773595in}{0.769629in}}%
\pgfpathlineto{\pgfqpoint{1.792810in}{0.762398in}}%
\pgfpathlineto{\pgfqpoint{1.812025in}{0.755167in}}%
\pgfpathlineto{\pgfqpoint{1.831239in}{0.747936in}}%
\pgfpathlineto{\pgfqpoint{1.850454in}{0.740705in}}%
\pgfpathlineto{\pgfqpoint{1.869669in}{0.733474in}}%
\pgfpathlineto{\pgfqpoint{1.888884in}{0.726243in}}%
\pgfpathlineto{\pgfqpoint{1.908099in}{0.719013in}}%
\pgfpathlineto{\pgfqpoint{1.927314in}{0.711782in}}%
\pgfpathlineto{\pgfqpoint{1.946529in}{0.704551in}}%
\pgfpathlineto{\pgfqpoint{1.965744in}{0.697320in}}%
\pgfpathlineto{\pgfqpoint{1.984958in}{0.690089in}}%
\pgfpathlineto{\pgfqpoint{2.004173in}{0.682858in}}%
\pgfpathlineto{\pgfqpoint{2.023388in}{0.675627in}}%
\pgfpathlineto{\pgfqpoint{2.042603in}{0.668396in}}%
\pgfpathlineto{\pgfqpoint{2.061818in}{0.661165in}}%
\pgfpathlineto{\pgfqpoint{2.081033in}{0.653934in}}%
\pgfpathlineto{\pgfqpoint{2.100248in}{0.646703in}}%
\pgfpathlineto{\pgfqpoint{2.119463in}{0.639472in}}%
\pgfpathlineto{\pgfqpoint{2.138677in}{0.632241in}}%
\pgfpathlineto{\pgfqpoint{2.157892in}{0.625011in}}%
\pgfpathlineto{\pgfqpoint{2.177107in}{0.617780in}}%
\pgfpathlineto{\pgfqpoint{2.196322in}{0.610549in}}%
\pgfpathlineto{\pgfqpoint{2.215537in}{0.603318in}}%
\pgfpathlineto{\pgfqpoint{2.234752in}{0.596087in}}%
\pgfpathlineto{\pgfqpoint{2.253967in}{0.588856in}}%
\pgfpathlineto{\pgfqpoint{2.273182in}{0.581625in}}%
\pgfpathlineto{\pgfqpoint{2.292397in}{0.574394in}}%
\pgfpathlineto{\pgfqpoint{2.311611in}{0.567163in}}%
\pgfpathlineto{\pgfqpoint{2.330826in}{0.559932in}}%
\pgfpathlineto{\pgfqpoint{2.350041in}{0.552702in}}%
\pgfpathlineto{\pgfqpoint{2.369256in}{0.545470in}}%
\pgfpathlineto{\pgfqpoint{2.388471in}{0.538240in}}%
\pgfpathlineto{\pgfqpoint{2.407686in}{0.531009in}}%
\pgfpathlineto{\pgfqpoint{2.426901in}{0.523778in}}%
\pgfpathlineto{\pgfqpoint{2.446116in}{0.516548in}}%
\pgfpathlineto{\pgfqpoint{2.465330in}{0.509318in}}%
\pgfusepath{stroke}%
\end{pgfscope}%
\begin{pgfscope}%
\pgfsetrectcap%
\pgfsetmiterjoin%
\pgfsetlinewidth{0.331237pt}%
\definecolor{currentstroke}{rgb}{0.150000,0.150000,0.150000}%
\pgfsetstrokecolor{currentstroke}%
\pgfsetdash{}{0pt}%
\pgfpathmoveto{\pgfqpoint{0.467944in}{0.349549in}}%
\pgfpathlineto{\pgfqpoint{0.467944in}{1.266874in}}%
\pgfusepath{stroke}%
\end{pgfscope}%
\begin{pgfscope}%
\pgfsetrectcap%
\pgfsetmiterjoin%
\pgfsetlinewidth{0.331237pt}%
\definecolor{currentstroke}{rgb}{0.150000,0.150000,0.150000}%
\pgfsetstrokecolor{currentstroke}%
\pgfsetdash{}{0pt}%
\pgfpathmoveto{\pgfqpoint{2.560444in}{0.349549in}}%
\pgfpathlineto{\pgfqpoint{2.560444in}{1.266874in}}%
\pgfusepath{stroke}%
\end{pgfscope}%
\begin{pgfscope}%
\pgfsetrectcap%
\pgfsetmiterjoin%
\pgfsetlinewidth{0.331237pt}%
\definecolor{currentstroke}{rgb}{0.150000,0.150000,0.150000}%
\pgfsetstrokecolor{currentstroke}%
\pgfsetdash{}{0pt}%
\pgfpathmoveto{\pgfqpoint{0.467944in}{0.349549in}}%
\pgfpathlineto{\pgfqpoint{2.560444in}{0.349549in}}%
\pgfusepath{stroke}%
\end{pgfscope}%
\begin{pgfscope}%
\pgfsetrectcap%
\pgfsetmiterjoin%
\pgfsetlinewidth{0.331237pt}%
\definecolor{currentstroke}{rgb}{0.150000,0.150000,0.150000}%
\pgfsetstrokecolor{currentstroke}%
\pgfsetdash{}{0pt}%
\pgfpathmoveto{\pgfqpoint{0.467944in}{1.266874in}}%
\pgfpathlineto{\pgfqpoint{2.560444in}{1.266874in}}%
\pgfusepath{stroke}%
\end{pgfscope}%
\begin{pgfscope}%
\pgfsetbuttcap%
\pgfsetmiterjoin%
\definecolor{currentfill}{rgb}{1.000000,1.000000,1.000000}%
\pgfsetfillcolor{currentfill}%
\pgfsetfillopacity{0.800000}%
\pgfsetlinewidth{0.331237pt}%
\definecolor{currentstroke}{rgb}{0.800000,0.800000,0.800000}%
\pgfsetstrokecolor{currentstroke}%
\pgfsetstrokeopacity{0.800000}%
\pgfsetdash{}{0pt}%
\pgfpathmoveto{\pgfqpoint{1.806147in}{0.863641in}}%
\pgfpathlineto{\pgfqpoint{2.482666in}{0.863641in}}%
\pgfpathquadraticcurveto{\pgfqpoint{2.504888in}{0.863641in}}{\pgfqpoint{2.504888in}{0.885863in}}%
\pgfpathlineto{\pgfqpoint{2.504888in}{1.189096in}}%
\pgfpathquadraticcurveto{\pgfqpoint{2.504888in}{1.211319in}}{\pgfqpoint{2.482666in}{1.211319in}}%
\pgfpathlineto{\pgfqpoint{1.806147in}{1.211319in}}%
\pgfpathquadraticcurveto{\pgfqpoint{1.783925in}{1.211319in}}{\pgfqpoint{1.783925in}{1.189096in}}%
\pgfpathlineto{\pgfqpoint{1.783925in}{0.885863in}}%
\pgfpathquadraticcurveto{\pgfqpoint{1.783925in}{0.863641in}}{\pgfqpoint{1.806147in}{0.863641in}}%
\pgfpathclose%
\pgfusepath{stroke,fill}%
\end{pgfscope}%
\begin{pgfscope}%
\pgfsetroundcap%
\pgfsetroundjoin%
\pgfsetlinewidth{0.501875pt}%
\definecolor{currentstroke}{rgb}{0.003922,0.450980,0.698039}%
\pgfsetstrokecolor{currentstroke}%
\pgfsetdash{}{0pt}%
\pgfpathmoveto{\pgfqpoint{1.828369in}{1.126228in}}%
\pgfpathlineto{\pgfqpoint{2.050592in}{1.126228in}}%
\pgfusepath{stroke}%
\end{pgfscope}%
\begin{pgfscope}%
\definecolor{textcolor}{rgb}{0.150000,0.150000,0.150000}%
\pgfsetstrokecolor{textcolor}%
\pgfsetfillcolor{textcolor}%
\pgftext[x=2.139480in,y=1.087339in,left,base]{\color{textcolor}\sffamily\fontsize{8.000000}{9.600000}\selectfont DNNR}%
\end{pgfscope}%
\begin{pgfscope}%
\pgfsetroundcap%
\pgfsetroundjoin%
\pgfsetlinewidth{0.501875pt}%
\definecolor{currentstroke}{rgb}{0.870588,0.560784,0.019608}%
\pgfsetstrokecolor{currentstroke}%
\pgfsetdash{}{0pt}%
\pgfpathmoveto{\pgfqpoint{1.828369in}{0.969055in}}%
\pgfpathlineto{\pgfqpoint{2.050592in}{0.969055in}}%
\pgfusepath{stroke}%
\end{pgfscope}%
\begin{pgfscope}%
\definecolor{textcolor}{rgb}{0.150000,0.150000,0.150000}%
\pgfsetstrokecolor{textcolor}%
\pgfsetfillcolor{textcolor}%
\pgftext[x=2.139480in,y=0.930167in,left,base]{\color{textcolor}\sffamily\fontsize{8.000000}{9.600000}\selectfont KNN}%
\end{pgfscope}%
\end{pgfpicture}%
\makeatother%
\endgroup%